%% file: main.tex
\documentclass[sigconf]{acmart}
\AtBeginDocument{%
  }

\setcopyright{acmlicensed}
\copyrightyear{2018}
\acmYear{2018}
\acmDOI{XXXXXXX.XXXXXXX}
\acmConference[Conference acronym 'XX]{Make sure to enter the correct
  conference title from your rights confirmation email}{June 03--05,
  2018}{Woodstock, NY}
\acmISBN{978-1-4503-XXXX-X/2018/06}

\usepackage{xcolor}   
\usepackage{xspace}

\usepackage{amsmath} 

\usepackage{algorithm}    
\usepackage{algpseudocode}  

\usepackage{booktabs}
\usepackage{multirow} 
\usepackage{tabularx} 
\usepackage{caption}  
\usepackage{enumitem}

\usepackage{color}
\usepackage{colortbl}
\definecolor{mygray}{gray}{.9}
\usepackage{amsmath}
\usepackage{stfloats}

\usepackage{listings}
\usepackage{tcolorbox}
\tcbuselibrary{skins,breakable,listings}

\newcommand{\method}{AgenticGEO}
\newcommand{\model}{{AgenticGEO}\xspace}

\newbool{showcomments}
\booltrue{showcomments}
\newcommand{\cmt}[3]{%
    \ifbool{showcomments}{%
        {\color{#1}\textbf{#2}: #3}%
    }{}%
}

\begin{document}

\title{AgenticGEO: A Self-Evolving Agentic System \\ for Generative Engine Optimization}

\author{Jiaqi Yuan}
\affiliation{%
  \institution{School of Computer Science and Engineering, Beihang University}
  \city{Beijing}
  \country{China}
  }
\email{yuanjq@buaa.edu.cn}

\author{Jialu Wang}

\affiliation{%
  \institution{Independent Contributor}
  \country{CA, United States}
  }
\email{faldict@ucsc.edu}

\author{Zihan Wang}
\affiliation{%
  \institution{School of Computer Science and Engineering, Beihang University}
  \city{Beijing}
  \country{China}
  }
\email{wzhan@buaa.edu.cn}

\author{Qingyun Sun}
\affiliation{%
  \institution{School of Computer Science and Engineering, Beihang University}
  \city{Beijing}
  \country{China}
  }
\email{sunqy@buaa.edu.cn}

\author{Ruijie Wang}
\authornote{Corresponding Author.}
\affiliation{%
  \institution{School of Computer Science and Engineering, Beihang University}
  \city{Beijing}
  \country{China}
  }
\email{ruijiew@buaa.edu.cn}

\author{Jianxin Li}
\affiliation{%
  \institution{School of Computer Science and Engineering, Beihang University}
  \city{Beijing}
  \country{China}
  }
  \email{lijx@buaa.edu.cn}

\renewcommand{\shortauthors}{Jiaqi Yuan, Jialu Wang et al.}

\begin{abstract}
Generative search engines represent a transition from traditional ranking-based retrieval to Large Language Model (LLM)-based synthesis, transforming optimization goals from ranking prominence towards content inclusion. Generative Engine Optimization (GEO), specifically, aims to maximize visibility and attribution in black-box summarized outputs by strategically manipulating source content. However, existing methods rely on static heuristics, single-prompt optimization, or engine preference rule distillation that is prone to overfitting. They cannot flexibly adapt to diverse content or the changing behaviors of generative engines. Moreover, effectively optimizing these strategies requires an impractical amount of interaction feedback from the engines. To address these challenges, we propose \model, a self-evolving agentic framework formulating optimization as a content-conditioned control problem, which enhances intrinsic content quality to robustly adapt to the unpredictable behaviors of black-box engines. Unlike fixed-strategy methods, \model~employs a MAP-Elites archive to evolve diverse, compositional strategies. To mitigate interaction costs, we introduce a Co-Evolving Critic, a lightweight surrogate that approximates engine feedback for content-specific strategy selection and refinement,
efficiently guiding both evolutionary search and inference-time planning. Through extensive in-domain and cross-domain experiments on two representative engines, \model~ achieves state-of-the-art performance and demonstrates robust transferability, outperforming 14 baselines across 3 datasets. Our code and model are available at: \url{https://github.com/AIcling/agentic_geo}.

\end{abstract}




\received{20 February 2007}

\maketitle


\input{1introduction}

\input{2relatedwork}

\input{3formulation}
\input{4method}
\input{5experiment}
\input{6conclusion}

\newpage
\bibliographystyle{ACM-Reference-Format}
\bibliography{ref}

\newpage

\appendix

\input{appendix_A}
\input{appendix_B}

\input{appendix_C}

\end{document}

%% file: 1introduction.tex
\section{Introduction}

\emph{Generative search engines} (e.g., Google AI Overviews~\cite{google_ai_overviews_ai_mode_2025,cai2025_google_ai_only_search_reuters}, Bing Search~\cite{bing_generative_search_blog_2024}, Perplexity AI~\cite{perplexity_help_how_it_works}) are increasingly dominant in information access, shifting users from browsing ranked webpages to consuming summarized answers directly provided by Large Language Models (LLMs). In contrast to traditional search engines that act as gateways to links, these systems retrieve evidence from multiple sources and compose it into a single, coherent summary, often accompanied by explicit citations~\cite{gao2023retrieval,menick2022teaching,nakano2021webgpt}. This paradigm shift fundamentally alters the web ecosystem, transforming the engine from a content ranker into a direct information summarizer.

This paper studies Generative Engine Optimization (GEO)~\cite{aggarwal2024geo,chen2025generative}, which is an emerging optimization problem induced by this transition. 
While traditional Search Engine Optimization (SEO)~\cite{shahzad2020new, almukhtar2021search} aims to maximize the position of a source content within a ranked list by optimizing retrieval signals (e.g., keywords and backlinks)~\cite{saeed2024exploring,ziakis2019important}, it is insufficient for modeling how LLMs synthesize and attribute evidence~\cite{nestaas2024adversarial,kumar2024manipulating}. In contrast, GEO targets two distinct objectives: (1) \emph{Visibility}, the extent to which a source's information is incorporated into the generated answer and (2) \emph{Attribution}, whether and where the source is explicitly cited. GEO is critical for the sustainability of the web ecosystem, as generative answers increasingly govern the allocation of user attention~\cite{brantner2025sourcing,google_ai_overviews_ai_mode_2025}.

\begin{figure}[t]
  \centering
  \includegraphics[width=0.9\linewidth]{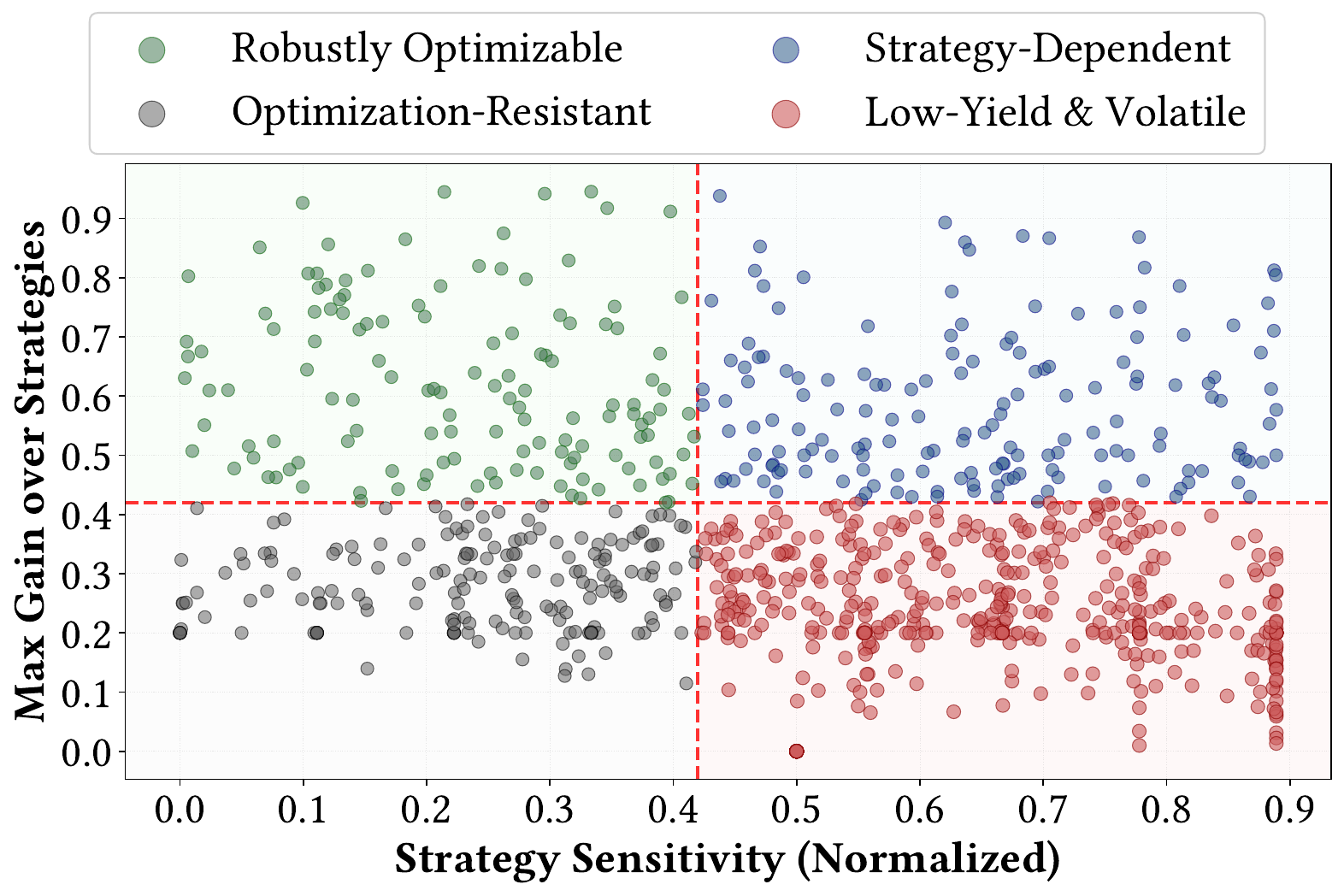}
 \caption{Characterization of the GEO result on GEO-Bench instances. $y$-axis reports maximum performance among 9 rewriting strategies, and $x$-axis reports performance variance among strategies. (i) Optimization success varies greatly by strategy and content. (ii) Existing strategies fail to optimize nearly half of instances (points in gray and red areas), indicating static strategy pool is not enough and needs evolving. Details can be found in Appendix~\ref{app:fig1}.}
 

  \label{fig:obs}
\end{figure}
Despite the growing interest in GEO, the field still remains under-explored. As evidenced by the strategy sensitivity analysis (Figure~\ref{fig:obs}), optimization success varies greatly by strategy and content. Meanwhile, existing strategies fail to optimize nearly half of the samples. These findings indicate the need for both customized strategy selection for each piece of content and refining the static strategy pool so it can adapt to new content patterns. However, existing work fail to achieve the goal, where they can be broadly categorized into: \textit{static heuristics approaches}~\cite{aggarwal2024geo} and \textit{learning-based approaches}~\cite{autogeo}. Static heuristic approaches apply heuristic rewriting strategies (i.e., rewriting prompt templates instructing an LLM) to source content.
However, this paradigm overlooks the heterogeneity of content and apply single strategy for all cotents. 
Learning-based approaches, in contrast, adapt rewriting strategies to the behavior of a specific generative engine (GE). Although effective in controlled settings, they tend to overfit to engine-specific patterns and degrade when the engine updates. In a non-stationary black-box environment, where retrieval, synthesis, and citation behaviors evolve over time, a static strategy pool is suboptimal and prone to miscalibration. Moreover, learning-based methods depend on frequent and intensive feedback from the specific generative engine during training, which is costly and often infeasible in real-world systems. These limitations highlight two key challenges for GEO: \emph{(i) Designing evolving methods that can flexibly adapt to diverse content and varying generative engine behaviors; (ii) Achieving effective optimization without relying on intensive feedback from generative engines.}

\begin{figure}[t]
  \centering
  \includegraphics[width=0.95\linewidth]{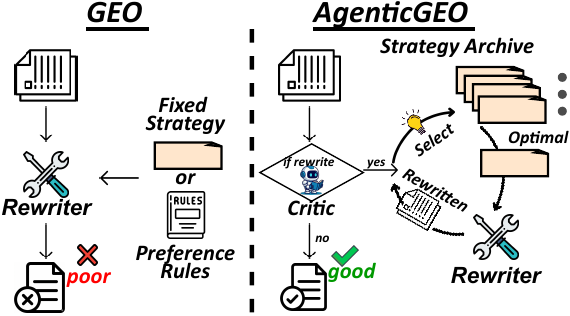}
  \caption{GEO v.s. AgenticGEO. Static GEO methods apply fixed rewriting heuristics, whereas AgenticGEO maintains an evolving strategy archive and a critic to adaptively retrieve high-scoring strategies for iterative rewriting.
  }
  \label{fig:comp}
\end{figure}

Motivated by these insights, we introduce \textbf{AgenticGEO}, a self-evolving agentic system that formulates GEO as learning a content-conditioned control policy, enhancing intrinsic quality for robust adaptation to black-box engines.
As illustrated in Fig\ref{fig:comp}, instead of applying a fixed rewrite heuristic, AgenticGEO maintains an evolving \emph{Quality-Diversity (QD) Archive} as external memory, preserving high-performing yet diverse strategies. 
Each strategy represents a distinct way to rewrite content under different structural, stylistic, or semantic preferences.
AgenticGEO further introduces a co-evolving critic to support agentic decision making. The critic serves as a surrogate evaluator and a planner. It identifies content-specific weaknesses, selects suitable strategies from the archive, and guides multi-step rewrites. By retrieving strategies from an evolved archive rather than relying on a fixed archive, AgenticGEO adapts naturally to diverse content and changing generative engine behaviors, addressing Challenge~(i).
To reduce reliance on intensive generative engine feedback, the critic is first calibrated using limited real feedback and then updated through continuous self-refinement. Once trained, it approximates generative engine preferences and provides stable guidance for strategy selection and rewrite execution. This design allows AgenticGEO to optimize visibility with substantially fewer feedback queries, addressing Challenge~(ii). Our analysis suggests archive-driven co-evolution admits a sublinear regret bound $O(\sqrt{T})$. Empirically, \model achieves the best optimization performance over 14 baselines on various benchmark datasets and generative engines ($46.4\%$ average gains. Moreover, \model manages to preserve $98.1\%$ performance using only $41.2\%$ sparse GE feedback for optimization, indicating that the evolving critic substantially reduces supervision reliance.

Our contributions are summarized as follows:
\begin{itemize}[leftmargin=*, nosep]
\item \textbf{Content-conditioned GEO formulation:}
Notably, we are the first to formulate GEO as a \emph{content-conditioned} optimization problem under non-stationary black-box generative engines, where different contents can favor different rewriting strategies.

\item \textbf{Co-evolving strategy memory and surrogate critic:}
We propose an agentic system that co-evolves a Quality-Diversity (QD) strategy archive as external memory and a lightweight surrogate critic that guides online exploration and inference-time multi-turn planning, enabling continual adaptation.

\item \textbf{Strong effectiveness and transfer:}
Extensive experiments show consistent improvements over baselines in-domain and strong transfer to unseen domains. Further analyses provide convergence evidence, validate the necessity of core component, and confirm that the critic serves as a reliable proxy for the generative engine, reducing reliance on expensive GE feedback.
\end{itemize}


%% file: 2relatedwork.tex
\section{Related Work}

\noindent\textbf{Generative Engine Optimization (GEO)}.
Online content optimization has traditionally focused on Search Engine Optimization ~\cite{shahzad2020new, almukhtar2021search, sharma2019brief, lewandowski2021seoInfluence}, 
which improves a page’s position in ranked Search Engine Results Pages (SERPs) by optimizing for ranking factors.
These factors typically combine retrieval-based relevance signals~\cite{manning2008ir} and link-analysis signals (e.g., PageRank~\cite{brin1998anatomy,page1999pagerank}), 
together with classic on-page/off-page heuristics such as keywords, metadata, and backlinks~\cite{saeed2024exploring,ziakis2019important,malaga2010search}.
With LLMs increasingly embedded into information access systems, user-facing search is shifting from ranked retrieval to retrieval-grounded answer synthesis in interactive, conversational settings~\cite{lewis2020retrieval,izacard2021leveraging,nakano2022webgpt}.

In the era of generative search, Aggarwal et al. introduced Generative Engine Optimization and released GEO-Bench~\cite{aggarwal2024geo}, 
reframing optimization as maximizing a source’s visibility within a generative engine’s synthesized response rather than competing for a rank position.
They show that lightweight rewriting edits (e.g., adding authoritative citations, inserting statistics, and crafting quotable statements) can substantially increase a source’s inclusion in GE outputs.
Building on this direction, AutoGEO~\cite{autogeo} distills engine preferences from LLM-generated explanations into rewriting rules, 
while RAID G-SEO~\cite{chen2025role} uses role-augmented intent inference and iterative reflection to guide intent-aligned rewriting.

Despite these advances, GEO remains in a developing stage. Most methods reduce GEO to LLM-based rewriting with fixed, hand-engineered prompts or static preference rules~\cite{aggarwal2024geo,autogeo,chen2025role}, which lack adaptability under black-box, dynamic engines and are sensitive to prompt formatting~\cite{sclar2024formatting}. 
Moreover, a specific strategy's effectiveness varies across domains and engines, yet existing methods rarely consider which rewriting strategy to apply based on the source content characteristics, limiting generalization and adaptation.

\noindent \textbf{Self-Evolving Agentic Systems}.
Self-evolving agents operate as closed-loop optimizers over system inputs, architectures, and environmental feedback, emerging as a key paradigm~\cite{agentgen,fang2025comprehensive,learnToRefine, mobilesteward,adaplanner,react,reflexion,selfrefine,voyager}. Existing systems are broadly organized as:

\noindent \underline{Policy Search.} \
Early works reframe prompts as discrete, optimizable variables: APE~\cite{zhou2022large} selects candidates via task-level scoring, while OPRO~\cite{yang2023large} iteratively proposes instructions based on prior scores.
However, these approaches often overfit to fixed protocols and lack online adaptivity.
Recent methods focus on inference-time adaptation. Self-Refine~\cite{selfrefine} and Reflexion~\cite{reflexion} iterate generation and feedback to revise solutions~\cite{yuksel2025multi}.
While effective for local errors, they typically follow fixed heuristic loops rather than learning to evolve, risking local optima when the generative engine updates.

\noindent \underline{Evolutionary Strategies.} \
To mitigate local optima, population-based algorithms like EvoPrompt and Promptbreeder~\cite{guo2023connecting,fernando2023promptbreeder} evolve prompts via LLM-based mutations and fitness selection.
Beyond prompts, recent systems extend this to agentic workflows~\cite{zhang2024aflow,wang2025evoagentx,zhang2025swarmagentic,zhang2025agentic,zhai2025agentevolver}.
This suggests that GEO requires a self-evolving architecture that updates task understanding and planning from black-box engine feedback, which remains under-explored.

%% file: 3formulation.tex
\section{Problem Formulation}

\label{sec:formulation}
\noindent\textbf{Black-box GEO setting}. 
We formulate Generative Engine Optimization as an optimization problem from the perspective of a content creator interacting with a black-box generative engine (GE), denoted as $\mathcal{E}$. Given a user query $q \in \mathcal{Q}$, the engine retrieves a candidate document set $D_q$ that includes the creator's original content $d$.
A rewriting strategy $s \in \mathcal{S}$ is applied to a policy (e.g., an LLM) to produce an optimized version $\tilde{d}$:
{
\begin{equation}
\tilde{d} = \mathrm{Rewrite}(d;\, s, q).
\end{equation}
}%
By substituting $d$ with $\tilde{d}$, the updated candidate set $\widetilde{D}_q = (D_q \setminus d) \cup \{\tilde{d}\}$ is processed by $\mathcal{E}$ to generate a synthesized response $\mathcal{A}$.

\noindent\textbf{Impression-based objective}.
The goal of the content creator is to identify an optimal strategy $s^\ast \in S$ that maximizes the visibility of the optimized content $\tilde{d}$ within the generated response $\mathcal{A}$.
Let $j^\star$ denote the rank index of $\tilde{d}$ in $\widetilde{D}$. Following the evaluation framework established in GEO-Bench~\cite{aggarwal2024geo}, we quantify the visibility using impression metrics that measure the presence and prominence of $\tilde{d}$ in $\mathcal{A}$. The optimization objective is formulated as:
{
\begin{equation}
\max_{s \in \mathcal{S}} \ \mathrm{Score}_{j^\star}(q, s),
\end{equation}
}%
where $\mathrm{Score}_{j^\star}(\cdot)$ represents a specific impression metric (e.g., \textsc{word}, \textsc{pos}, or \textsc{overall} impression; see Appendix~\ref{app:metrics} for definitions).

%
%
%
%

%% file: 4method.tex
\section{AgenticGEO Methodology}

\subsection{Overview}
\begin{figure*}[t]
  \centering  \includegraphics[width=1.0\textwidth]{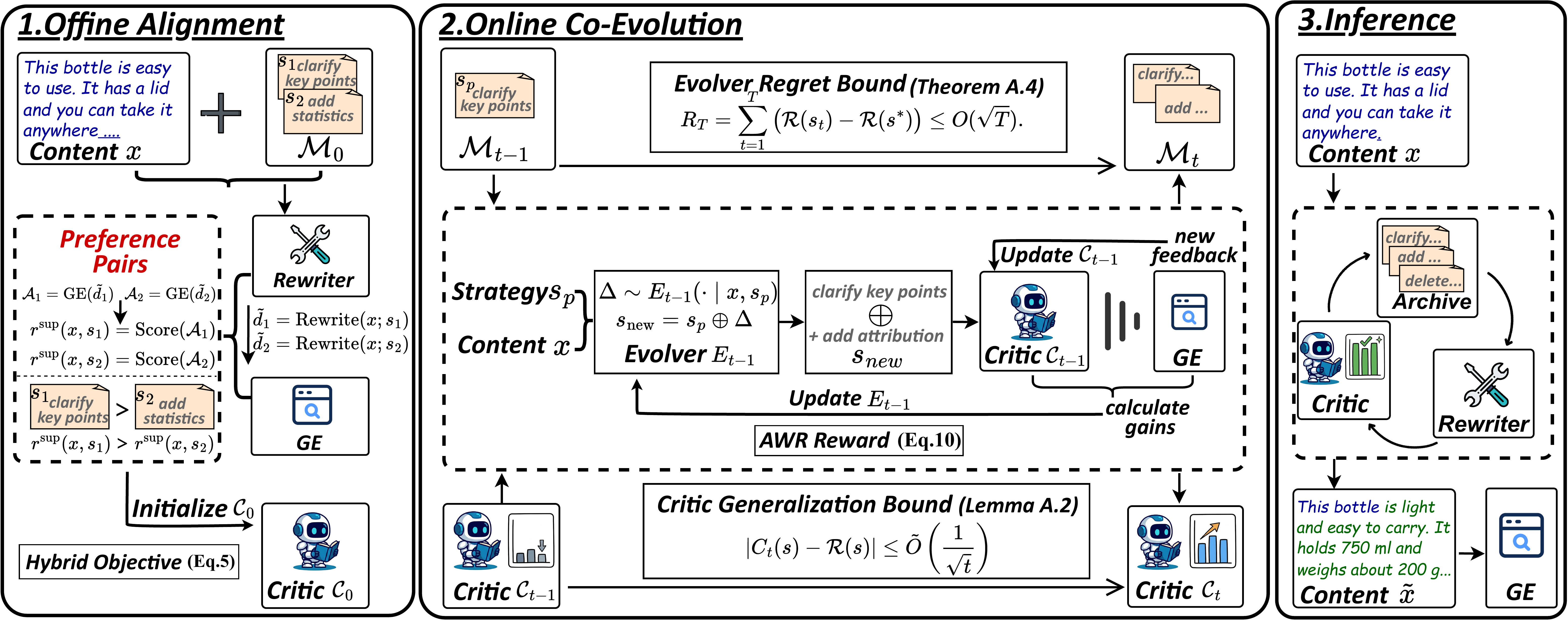} 

  \caption{Overview of the AgenticGEO framework with two-stage training. Offline Alignment warm-starts a surrogate critic using offline preference pairs from the initial archive $\mathcal{M}_0$, calibrating it for fast, content-conditioned strategy scoring. Online Co-Evolution then interacts with the black-box Generative Engine (GE) to iteratively, simultaneously evolve a MAP-Elites quality-diversity archive and the critic. Parent strategies are mutated by a learned evolver trained with sibling-aware AWR, the critic screens candidates to reduce GE calls, and newly collected GE feedback is stored in a replay buffer to continually recalibrate the critic and update the archive through a value-novelty gate. At inference time, the evolved archive and critic enable agentic multi-turn rewriting by selecting and executing a content-adaptive plan of strategies. 
  }
  \label{fig:overview}
\end{figure*}

As Figure~\ref{fig:overview} shows, AgenticGEO proceeds in three stages:

\begin{itemize}[leftmargin = 8pt]
    \item \textbf{Offline Critic Alignment:} We warm-start a lightweight surrogate \emph{critic} using offline preference pairs from the training dataset to approximate GE feedback, without costly online evaluations.

    \item \textbf{Online Co-Evolution:} Through a co-evolutionary loop, we jointly train the MAP-Elites strategy archive and the critic module with the real GE interactions.

    \item \textbf{Agentic Multi-Turn Rewriting:} At inference time, we perform agentic multi-step planning, where the critic orchestrates strategy selection, while the rewriter operates the chosen strategies to optimize content.
\end{itemize}

\vspace{-1.5ex}
\subsection{Offline Critic Preference Alignment}
To avoid the high latency of online interactions with the black-box engine, we train a lightweight \emph{critic} to serve as a surrogate evaluator. This critic is aligned with offline engine feedback to learn strategy-conditioned preferences, enabling an efficient warm-start.

\noindent \textbf{Setup \& Notation.}
Given a query $q$, a document $d$, and a rewriting strategy $s \in \mathcal{M}_0$ (see Appendix~\ref{app:seed}), we denote the input context as $x=(q,d)$.
Each strategy $s$ is instantiated as a textual prompt template that instructs an LLM to rewrite source content $d$.

\noindent \textbf{Architecture \& Context Encoding.}
We implement the critic using a \emph{backbone + value head} structure, denoted by $\mathcal{C}$.
We select a lightweight decoder-only Language Model (LM) as the backbone to leverage its inherent semantic reasoning capabilities, which are essential for capturing the complex dependencies between optimization strategies and the query-content context.

Given context $x$ and strategy $s$, the backbone encodes their concatenation into a latent representation $h(x,s)$, which is projected by a two-layer MLP value head to a numerical score:
\begin{equation}
h(x,s)=\mathrm{LM}\!\left([x; s]\right),\qquad 
\mathcal{C}(x,s)=\mathrm{MLP}\!\Big(h(x,s)\Big),
\end{equation}
where $\mathcal{C}(x,s)$ is expected to predict the impression gain induced by applying strategy $s$ to context $x$ before feeding it into GE.

\noindent \textbf{Offline Supervision Data Construction.}
We construct offline supervision from the seed strategy pool $\mathcal{M}_0$ (illustrated at Appendix~\ref{app:seed}).
For each context $x$ and strategy $s$, we define the supervised gain as the improvement over the unrewritten baseline:
\begin{equation}
r^{\text{sup}}(x,s)=\mathrm{Score}\!\left(\mathcal{A}^{\text{train}}_{s}\right)-\mathrm{Score}\!\left(\mathcal{A}^{\text{train}}_{0}\right),
\end{equation}
where $\mathcal{A}^{\text{train}}_{s}$ denotes the generative engine output after applying strategy $s$ to $x$, and $\mathcal{A}^{\text{train}}_{0}$ is the corresponding unrewritten baseline output. $\mathrm{Score}(\cdot)$ is the \textsc{overall} impression metric defined in Appendix~\ref{app:metrics}, combining \textsc{Word} and \textsc{Pos}. We use $r^{\text{sup}}(x,s)$ as the offline alignment target for the critic output $\mathcal{C}(x,s)$.

\noindent \textbf{Hybrid Objective.}
Effective preference alignment requires capturing both the \emph{absolute value} and the \emph{relative order} of strategies. We propose a hybrid objective combining regression and ranking:
\begin{equation}
\label{eq:hybrid}
\mathcal{L}_{\text{total}}=\mathcal{L}_{\text{pair}}+\lambda\mathcal{L}_{\text{reg}}.
\end{equation}

\textbf{(1) Score Regression:}
We use the Huber loss~\cite{huber1992robust} to regress $\mathcal{C}(x,s)$ onto $r^{\text{sup}}(x,s)$, which is less sensitive to noisy supervision:
\begin{equation}
\mathcal{L}_{\text{reg}}=\mathbb{E}_{(x,s)}\Big[\textsc{Huber}\big(\mathcal{C}(x,s),\, r^{\text{sup}}(x,s)\big)\Big].
\end{equation}

\textbf{(2) Rank-Aware Pairwise Alignment:}
While regression calibrates the value scale, downstream strategy selection primarily depends on relative ordering.
We therefore construct pairwise strategy preferences within the same context:
for each $x$, we rank strategies in $\mathcal{M}_0$ by $r^{\text{sup}}(x,s)$ (rank $1$ is best), and sample ordered pairs $(s^+,s^-)$ such that
$r^{\text{sup}}(x,s^+) > r^{\text{sup}}(x,s^-)$.
Since accurate discrimination among top strategies is most important for strategy selection, we assign larger weights to pairs involving higher-ranked strategies:
\begin{equation}
w(s^+,s^-)=\frac{1}{\text{rank}_x(s^{+})+\text{rank}_x(s^{-})}.
\end{equation}
The weighted pairwise loss emphasizes the most promising strategies for reliable selection:
\begin{equation}
\hspace{-2ex}
\mathcal{L}_{\text{pair}}
=\mathbb{E}_{(x,s^{+},s^{-})}\!\left[
w(s^+,s^-)\cdot \log\!\Big(1+e^{-(\mathcal{C}(x,s^{+})-\mathcal{C}(x,s^{-}))}\Big)
\right].
\end{equation}

\noindent \textbf{Staged Training Strategy.}
To further stabilize alignment, we employ a two-phase process. We primarily sample \emph{Top-$5$ dense pairs} to refine fine-grained local ordering, and \emph{global contrastive pairs} to ensure coarse separation. We initially freeze the backbone to warm up the value head, preventing representation collapse, before unfreezing all parameters for joint fine-tuning.

\subsection{Online Strategy-Critic Co-Evolution}
\label{sec:online_evolution}

While offline alignment initializes the system, relying on static strategies risks local optima and fails to adapt to dynamic search environments. To enable continuous adaptation, we introduce an \emph{Online Strategy--Critic Co-Evolution} framework. This establishes a self-evolving loop where the \emph{Evolver} ($E$), a parameterized LLM that generates strategy mutations, actively expands the strategy space to discover novel ones, while the \emph{Critic} ($\mathcal{C}$) continuously recalibrates to guide exploration and enable optimal strategy selection at inference.

\input{Tables/genotype}

\subsubsection{Structured Evolution via MAP-Elites Archive}
\label{sec:map_elites_archive}
To prevent the optimizer from collapsing into a single ``safe'' pattern (e.g., always using an authoritative tone) that fails on diverse content, we maintain a dynamic \emph{MAP-Elites Archive} $\mathcal{M}$~\cite{mouret2015illuminating,pugh2016quality,justesen2019map}. Instead of seeking one global optimum, this archive acts as an evolving memory that preserves a wide range of high-performing strategies.

Unlike a standard top-$k$ list that discards lower-scoring but distinct solutions, $\mathcal{M}$ organizes strategies into a multi-dimensional grid of \emph{behavioral cells}. Each cell represents a specific combination of attributes (see Table~\ref{tab:gen}), such as an Assertive tone combined with a List format, ensuring that unique strategy styles compete only against similar ones. A new strategy $s$ captures a cell only if it triggers the \emph{Value-Novelty Gate} (details in Appendix~\ref{subsec:app_gates}):
\begin{enumerate}[leftmargin=12pt]
    \item \textbf{Value:} It achieves a higher impression score from the Generative Engine than the current elite in that cell;
    \item \textbf{Novelty:} It is structurally distinct from existing entries (measured by $n$-gram distance~\cite{broder1997resemblance}), expanding the archive's coverage even if its score is currently lower.
\end{enumerate}

To manage archive capacity and support evolution exploration, we assign each retained strategy a composite \emph{PND Score} (Pareto-Novelty-Diversity)~\cite{lehman2011evolving,deb2002fast}:
\begin{equation}
S_{\mathrm{PND}}(s) = r(s) + \lambda_{\mathrm{pnd}} \cdot \big(\operatorname{Nov}(s) + \operatorname{Div}(s)\big),
\label{eq:pnd}
\end{equation}
where $r(s)$ is the impression score from the critic or generative engine, and $\operatorname{Nov}(s)$ and $\operatorname{Div}(s)$ measure structural uniqueness and lineage diversity (see Appendix~\ref{subsec:app_pnd}). This score serves two roles: \emph{Global Pruning} to discard redundant strategies when the archive is full, and a \emph{dense intrinsic reward} for the Evolver (Eq.~\ref{eq:sibling_adv}) to encourage exploration beyond pure exploitation.

\begin{algorithm}[t]
\caption{Archive-Driven Strategy-Critic Co-Evolution}
\label{alg:co_evolution}
\begin{algorithmic}[1]
\Require Initial archive $\mathcal{M}_0$; data distribution $\mathcal{D}$ over content;
evolver $E$; critic $\mathcal{C}$; generative engine $\mathrm{GE}$;
online iterations $T$; exploit size $K_{\text{top}}$; explore size $K_{\text{rand}}$.
\State $\mathcal{M} \leftarrow \mathcal{M}_0$
\State Initialize replay buffers $\mathcal{B}_{\text{true}} \leftarrow \emptyset,\ \mathcal{B}_{\text{pred}} \leftarrow \emptyset$
\For{$t = 1,\dots,T$}
    \State $x \sim \mathcal{D}$ \Comment{e.g., $(q,d)$}
    \State \textbf{Phase 1: Hybrid Candidate Generation}
    \State $P \leftarrow \mathrm{Sample}(\mathcal{M}_t)$
    \State $S_{\text{evolver}} \leftarrow \{\, s \mid s \sim E(\cdot \mid s_p),\ s_p \in P \,\}$ \Comment{neural mutation}
    \State $S_{\text{ops}} \leftarrow \{\, \mathrm{Mutate}(s_p) \mid s_p \in P \,\}$ \Comment{symbolic perturbation}
    \State $S_{\text{cand}} \leftarrow S_{\text{evolver}} \cup S_{\text{ops}}$
    \State \textbf{Phase 2: Critic Scoring \& Budgeted Selection}
    \State $R_{\text{critic}}(s) \leftarrow \mathcal{C}(x,s),\ \forall s \in S_{\text{cand}}$
    \State $S_{\text{eval}} \leftarrow \mathrm{TopK}(S_{\text{cand}}, R_{\text{critic}}, K_{\text{top}})\ \cup\ \mathrm{Random}(S_{\text{cand}}, K_{\text{rand}})$
    \State \textbf{Phase 3: GE Evaluation \& Joint Reward Aggregation}
    \State $R_{\text{true}}(s) \leftarrow \mathrm{GE}(x,s),\ \forall s \in S_{\text{eval}}$ 
    \State $R_{\text{mix}}(s) \leftarrow 
    \begin{cases}
        R_{\text{true}}(s), & s \in S_{\text{eval}},\\
        R_{\text{critic}}(s), & s \in S_{\text{cand}} \setminus S_{\text{eval}}.
    \end{cases}$
    \State $\mathcal{M}_{t+1} \leftarrow \mathrm{UpdateArchive}(\mathcal{M}_{t}, S_{\text{cand}}, R_{\text{mix}})$
    \State $\mathcal{B}_{\text{true}} \leftarrow \mathcal{B}_{\text{true}} \cup \{(x,s,R_{\text{true}}(s)) \mid s \in S_{\text{eval}}\}$
    \State $\mathcal{B}_{\text{pred}} \leftarrow \mathcal{B}_{\text{pred}} \cup \{(x,s,R_{\text{mix}}(s)) \mid s \in S_{\text{cand}}\setminus S_{\text{eval}}\}$
    \State \textbf{Phase 4: Online Updates}
    \State $E_{t+1} \leftarrow \mathrm{TrainEvolver}(E_{t}, \mathcal{B}_{\text{true}} \cup \mathcal{B}_{\text{pred}})$
    \State $\mathcal{C}_{t+1} \leftarrow \mathrm{TrainCritic}(\mathcal{C}_{t}, \mathcal{B}_{\text{true}})$
\EndFor
\State \Return Evolved archive $\mathcal{M}$, evolved critic $\mathcal{C}$
\end{algorithmic}
\end{algorithm}

\subsubsection{The Co-Evolutionary Loop}
With the diverse population anchored by the Archive, the online process drives a \emph{co-evolutionary loop} where the Evolver and Critic mutually refine their capabilities through four phases per iteration (shown in Algorithm~\ref{alg:co_evolution}):
\begin{enumerate}[label=\arabic*.,leftmargin=*]
\item \textbf{Generation:} Parents sampled from $\mathcal{M}$ undergo hybrid mutation. The evolver $E$ selects an operator from our predefined catalog and generates the resulting \texttt{child\_genotype} (e.g., applying \texttt{mut\_F\_schema\_swap} to change the output format), while symbolic Operators inject hard perturbations via field-level mutations (e.g., \texttt{mut\_T\_toggle\_tone} for style switching, or \texttt{mut\_C\_st-
rengthen} for constraint injection). Details in Appendix~\ref{subsec:app_evolver}. 

\item \textbf{Screening:} To reduce computational cost, the \emph{Critic} $\mathcal{C}$ filters candidates, selecting the Top-$K_{\text{top}}$ strategies for exploitation and a random $K_{\text{rand}}$ subset for exploration to mitigate selection bias.

\item \textbf{Evaluation:} The Generative Engine evaluates the selected candidates. The resulting GE feedback, together with the critic scores for the remaining candidates, is merged into a joint reward signal to update the archive via the Value-Novelty gate, and all experiences are logged into the replay buffers $\mathcal{B}_{\text{true}}$ and $\mathcal{B}_{\text{pred}}$ for subsequent online updates.

\item \textbf{Learning:} The replay buffers drive online updates of the evolver $E$ and recalibration of the critic $C$. $E$ is trained on both $\mathcal{B}_{\text{true}}$ and $\mathcal{B}_{\text{pred}}$, while $C$ is updated only with GE-labeled samples in $\mathcal{B}_{\text{true}}$, enabling continual adaptation as the strategy population evolves.
\end{enumerate}

\subsubsection{Evolver Optimization via Sibling-Aware AWR}
The Evolver $E$ synthesizes a candidate strategy $s_{\text{new}}$ from a parent $s_{p}$ (or a parent pair) by selecting optimal mutation or crossover operators. A naive Reinforcement Learning approach is unstable due to the high variance of impression scores from the Generative Engine. We instead employ \emph{Advantage-Weighted Regression (AWR)}.

Inspired by GRPO-style relative advantage calculation~\cite{shao2024deepseekmath}, we further stabilize learning under noisy feedback.
Crucially, to mitigate noise across heterogeneous content contexts (where some source material is inherently harder to optimize), we propose a \emph{Sibling-Aware Advantage}.
Instead of comparing rewards globally, we compare a candidate's performance relative to its ``siblings'' generated from the same parent strategy:
\begin{equation}
A_i
= \underbrace{(r_i - r_{\mathrm{parent}})}_{\text{Absolute Gain}}
- \alpha_{\mathrm{sib}} \cdot \operatorname{mean}\!\left(\{\Delta_j\}_{j \in \mathrm{siblings}}\right)
+ \mathbb{I}(\Delta_i < 0)\cdot S_{\mathrm{PND}}(s_i),
\label{eq:sibling_adv}
\end{equation}
where $r$ denotes the score from the critic and generative engine, $\Delta_i = r_i - r_{\mathrm{parent}}$, and $S_{\mathrm{PND}}(s_i)$ is the exploration bonus (Eq.~\ref{eq:pnd}).
The sibling mean provides a within-parent baseline, so $A_i$ better reflects the effect of the chosen evolution operator rather than the intrinsic difficulty of the content.
The last term adds an exploration bonus when the immediate gain is non-positive, helping retain novel and diverse strategies.
The policy is then updated to imitate these high-advantage actions through weighted supervised fine-tuning (SFT)~\cite{ouyang2022training}, minimizing the following loss:
\begin{equation}
\mathcal{L}_{\mathrm{Evolver}} = -\mathbb{E}_{(x, s) \sim (\mathcal{B}_{\mathrm{true}} \cup \mathcal{B}_{\mathrm{pred}})} \left[ \exp\left(\frac{A(x,s)}{\beta}\right) \cdot \log E(s|x) \right].
\label{eq:awr}
\end{equation}

\subsubsection{Online Critic Calibration}
Complementing the Evolver updates, we continuously recalibrate the critic $\mathcal{C}$ using new labeled triplets $(x, s, r)$ collected in the replay buffer $\mathcal{B}_{\mathrm{true}}$. By optimizing the hybrid objective in Eq.~\ref{eq:hybrid}, we keep the critic calibrated under the evolving archive and feedback distribution, enabling reliable scoring for online selection and inference-time agentic planning.

\subsection{Theoretical analysis.}
Let $\mathcal{R}(s)$ denote the risk induced by the strategy $s$. The regret of the proposed critic-evolver co-evolutionary algorithm is analyzed in Theorem~\ref{thm:informal}.
\begin{theorem}[Informal]\label{thm:informal}
Under the conditions of a linearly growing replay buffer and a Lipschitz-continuous critic, the AgenticGEO framework achieves a cumulative regret:
\[
    \sum_{t=1}^T \mathcal{R}(s_t) - \mathcal{R}(s^*) = \mathcal{O}(\sqrt{T}).
\]
\begin{proof}[Proof Sketch]
We first decompose the instantaneous risk gap at each time step $t$ as
    \begin{multline*}
        \mathcal{R}(s_t) - \mathcal{R}(s^\ast) 
        \leq |\mathcal{R}(s_t) - \mathcal{C}_t(s_t)| + |\mathcal{C}_t(s_t) - \mathcal{C}_t(s^\ast)| + |\mathcal{C}_t(s^\ast) - \mathcal{R}(s^\ast)|.
    \end{multline*}
Given a replay buffer that grows linearly, the approximation and generalization errors of the critic model satisfy
\(
    |\mathcal{R}(s) - \mathcal{C}_t(s)| = \mathcal{O}(\frac{1}{\sqrt{t}}).
\)
The evolver's selection process follows standard online learning bound by
\(
    |\mathcal{C}_t(s_t) - \mathcal{C}_t(s^\ast)| = \mathcal{O}(\frac{1}{\sqrt{t}}).
\)
Combining above and summing over the time horizon $T$, we can conclude that the cumulative regret is
\[
    \sum_{t=1}^T \mathcal{R}(s_t) - \mathcal{R}(s^\ast)  = \sum_{t=1}^T \mathcal{O}(\frac{1}{\sqrt{t}}) = \mathcal{O}(\sqrt{T}).
\]
\end{proof}

\end{theorem}
This implies that as the number of iterations $T$ increases, the average performance gap vanishes, guaranteeing that the system asymptotically converges to the optimal strategy. More detailed theoretical analysis and the proofs are provided in Appendix~\ref{appendix:theorem}.

\subsection{Agentic Multi-Turn Rewriting with Critic-Guided Planning}

The inference phase deploys the evolved archive $\mathcal{M}$ and \textbf{critic} $\mathcal{C}$ for multi-turn optimization. 
We formulate this as a \textit{content-conditioned decision-making} process, where the critic serves as a \textbf{fast proxy planner}. 
This allows the agent to guide a greedy search over the strategy space, enabling rapid evaluation while avoiding expensive black-box interactions.

At each step $\tau$ (initialized with $d_0=d$), the agent plans the next optimal move by selecting a strategy $s_\tau^*$ that maximizes the critic's predicted potential, while enforcing a Tabu List $\mathcal{T}_\tau$ (recording previously utilized strategies) to prevent repeated operations:
\begin{equation}
    s_\tau^* = \operatorname*{arg\,max}_{s \in \mathcal{M} \setminus \mathcal{T}_\tau} \; \mathcal{C}\big((q, d_\tau), s\big).
\end{equation}

Subsequently, the content state transitions via the rewriting tool, and the utilized strategy is recorded to update the constraint set:
\begin{equation}
    d_{\tau+1} = \operatorname{Rewrite}(d_\tau, s_\tau^*, q), \quad \mathcal{T}_{\tau+1} \leftarrow \mathcal{T}_\tau \cup \{s_\tau^*\}.
\end{equation}

The loop terminates when the marginal gain vanishes:
\begin{equation}
\max_{s \in \mathcal{M} \setminus \mathcal{T}_{\tau+1}} \mathcal{C}\big((q, d_{\tau+1}), s\big)
\le
\max_{s \in \mathcal{M} \setminus \mathcal{T}_{\tau}} \mathcal{C}\big((q, d_{\tau}), s\big),
\end{equation}
or steps exceed $T_{max}$.
This agentic planning capability allows \method{} to dynamically adapt its optimization path based on the evolving characteristics of the content, yielding a highly optimized content that occupies a more prominent role within the engine's information synthesis and attribution.

%% file: Tables/genotype.tex
\begin{table}[t]
  \centering
  \setlength{\tabcolsep}{6pt} 
  \caption{Structure of the evolving strategy representation. Each dimension is mutated independently. It can be rendered into a compact summary for critic scoring and a full strategy prompt for rewriting (see Appendix~\ref{subsec:app_genotype}).}
  \label{tab:gen}
  \begin{tabularx}{\linewidth}{@{}l X@{}}
    \toprule
    \textbf{Dimension} & \textbf{Semantics \& Examples} \\
    \midrule
    Instruction & 
    Defines goal and scope (e.g., target audience, core facts, key emphasis, expert role). \\
    Constraints & 
    Sets strict boundaries (e.g., word count, citation checks, anti-hallucination, fact consistency). \\
    Reasoning & 
    Adds logic steps (e.g., conflict resolution, self-correction, step planning, logic verification). \\
    Format & 
    Controls output layout (e.g., bullet lists, code blocks, output schema, section preludes). \\
    Tone & 
    Adjusts writing style (e.g., assertive voice, technicality, simple language, formality level). \\
    \bottomrule
  \end{tabularx}
\end{table}

%% file: 5experiment.tex
\section{Experiments}


We empirically validate the effectiveness and generalization ability of AgenticGEO. We study the following research questions:
\begin{enumerate}[label=\textbf{RQ\arabic*}, leftmargin = *, wide=0pt]
    \item \textbf{Overall Performance \& Robustness:} How does AgenticGEO compare to state-of-the-art methods, and is its performance robust to generative engines varying in architecture and scale?
    
    \item \textbf{Transferability to Unseen Domains:} Can an optimization policy maintain performance when deployed on out-of-distribution domains?
    
    \item \textbf{Ablation and Hyperparameters Analysis:} How does each co-evolutionary component influence the performance? Specifically, does the pre-trained critic provide a reliable warm-start?


    \item \textbf{Semantic Consistency:} Does the optimization maintain the original meaning of the content, ensuring that visibility gains do not come at the cost of information loss?
\end{enumerate}

\subsection{Exprimental Setup}
We briefly introduce the experimental setup.
\begin{itemize}[leftmargin=*, nosep]
    \item \textbf{Dataset.} \ GEO-Bench~\cite{aggarwal2024geo} serves as our training dataset, derived from Google Search results spanning a wide spectrum of domains and query difficulties. To assess zero-shot generalization to unseen distributions, we employ MS MARCO~\cite{nguyen2016ms}, comprising short-text passages from real-world Bing search logs, and a custom E-commerce~\cite{reddy2022shopping} sourced from Amazon, representing a specific vertical for product search. Details are in Appendix~\ref{app:dataset}
    \item \textbf{Settings.} To conduct a comprehensive evaluation, we use GEO-Bench~\cite{aggarwal2024geo} training dataset for evolving the archive and critic, and extend the assessment to MS-Marco~\cite{nguyen2016ms} and E-commerce~\cite{reddy2022shopping} datasets. This diverse benchmark allows us to examine the performance consistency across varying content distributions and verify its broad applicability in real-world scenarios.
    \item \textbf{Baselines.} 
    We group baselines into two categories:
    Static heuristics apply fixed rewriting heuristics: 
    \textit{No optimization}, \textit{Keyword Stuffing}, \textit{Unique Words}, \textit{Easy-To-Understand}, \textit{Authoritative}, \textit{Technical Words}, \textit{Fluency Optimization}, \textit{Cite Sources}, \textit{Quotation Addition}, \textit{Statistics Addition}.
    Learning-based methods train models to generate optimized rewrites:
    \textit{AutoGEO}, \textit{Cite Sources-SFT}, \textit{Quotation Addition-SFT}, \textit{Statistics Addition-SFT}.
    Details are deferred to Appendix~\ref{app:baseline}.
    \item \textbf{Models \& Metrics.} \  
    We employ Qwen2.5-32B-Instruct and Llama-3.3-70B-Instruct as downstream generative engines~\cite{qwen2025qwen25technicalreport,grattafiori2024llama3herdmodels}. For system components, we implement the Critic backbone with Qwen2.5-1.5B, the Evolver with Qwen2.5-7B-Instruct, and the Rewriter with Qwen2.5-32B-Instruct for tool invocation. We measure performance via \textit{Attributed \textbf{Word} Count}, \textit{\textbf{Pos}ition-Weighted Citation Order}, and their Combination as \textit{\textbf{overall}}.~\cite{aggarwal2024geo} (Definition in Appendix~\ref{app:metrics}.)
    \item \textbf{Implementation.} \ We employ LoRA~\cite{hu2022lora} to fine-tune both the Critic and Evolver for 2 epochs. All experiments are implemented on 4 NVIDIA RTX Pro 6000 GPUs. At inference time, we select the top-$25$ strategies from the evolved archive ranked by their $S_{\mathrm{PND}}$ scores, and perform critic-guided multi-turn rewriting with a maximum of $3$ rewrite steps. Other details in Appendix~\ref{app:exp}.
\end{itemize}




\vspace{-2ex}
\subsection{Overall Performance and Robustness (RQ1).}
Table~\ref{tab:exp/geo-bench} presents the comparative results of AgenticGEO against all baselines on the in-domain GEO-Bench dataset.
We observe that AgenticGEO consistently achieves state-of-the-art performance on two generative engines, demonstrating strong effectiveness and robustness against engine variations in architecture and scale.

On the Qwen2.5-32B-Instruct engine, AgenticGEO achieves an Overall score of 25.48, surpassing the strongest baseline (AutoGEO) which scores 23.71. This represents a substantial improvement over static heuristic strategies, such as \textit{Keyword Stuffing} (20.69) and \textit{Authoritative} (20.60), confirming that fixed rewriting rules are insufficient for the dynamic nature of generative engines.
Furthermore, our method outperforms the Supervised Fine-Tuning (SFT) baselines. This performance improvement indicates that our self-evolving strategy archive effectively transcends the optimization upper bound imposed by static strategies' supervision, capturing content-centric patterns that are ignored by existing methods.

AgenticGEO also demonstrates strong robustness on the larger Llama-3.3-70B-Instruct engine.
While many baselines struggle to transfer their gains (e.g., \textit{Statistics Addition} drops to 21.05), AgenticGEO maintains a strongest performance with Overall of 24.52.
This consistency across different model architectures and scales validates that our co-evolving critic and strategy archive learn generalized optimization principles rather than overfitting to a specific engine, presenting the necessity of keeping strategy diverse.

\input{Tables/exp1}
\input{Tables/exp2}

\vspace{-1ex}
\subsection{Transferability to Unseen Domains (RQ2).}
To evaluate AgenticGEO's cross-domain transferability, we test AgenticGEO on MS MARCO  and E-Commerce without domain-specific fine-tuning.
As shown in Table~\ref{tab:exp/cross-domain}, our method exhibits strong robustness against domain shifts, whereas baselines suffer from significant performance degradation. On MS MARCO, AgenticGEO outperforms the strongest baseline, AutoGEO, by over 11\% on both Qwen2.5-32B-Instruct and Llama3.3-70B-Instruct. And the advantage is even more dominant on the E-Commerce dataset.

The transferability gains across two unseen domains support the claim that AgenticGEO avoids overfitting to specific content. 
The design of an evolving strategy archive and critic-guided planning yields a transferable optimization policy that remains effective under domain shift, showing strong domain generalization.






\vspace{-1ex}
\subsection{Ablation Analysis (RQ3)}

\begin{figure*}[t] 
  \centering  \includegraphics[width=0.95\textwidth]{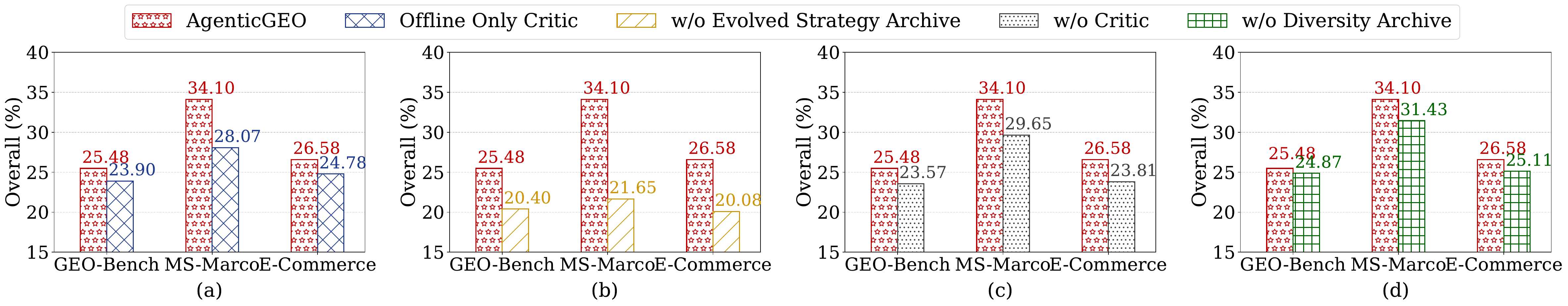} 
    \vspace{-3ex}
\caption{Ablation study of AgenticGEO on three datasets. We compare AgenticGEO with four variants: (a) an offline-only critic trained without online co-evolution, (b) removing the evolved strategy archive, (c) replacing the critic with random rewrite planning and the evolver directly generates the strategy, (d) maintaining the archive by performance only without diversity.}
  \label{fig:ablation_study}
\end{figure*}

\paragraph{Impact of Core Components.}
Figure~\ref{fig:ablation_study} shows that removing any of the components degrades performance on all datasets. 
The largest drop comes from removing the evolved strategy archive (b),  confirming that long-term strategy accumulation is the primary driver of gains. 
Using an offline-only critic (a) is also clearly weaker, highlighting the importance of online co-evolution and continual critic recalibration. Replacing critic-guided planning with random planning (c) and maintaining the archive by performance only (d) cause degrades, suggesting diversity-aware archive improves generability. 

\begin{figure}[t]
    \centering
    \includegraphics[width=\columnwidth]{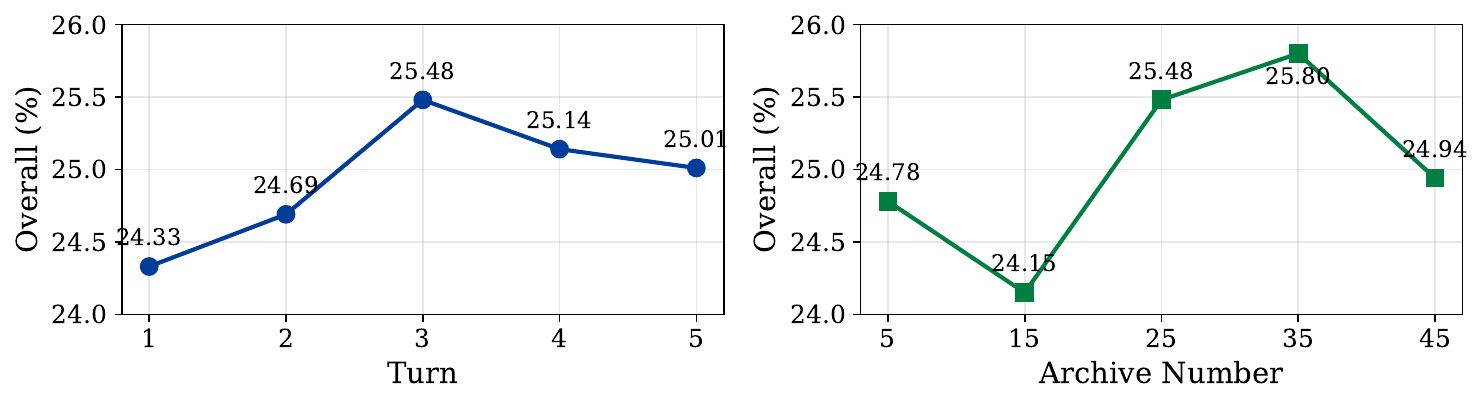}
    \caption{Hyper-parameter sensitivity analysis on GEO-Bench. We report overall impression score when varying the multi-turn rewriting steps (left) and the archive size (right).}
    \label{fig:exp/sensitivity}
\end{figure}

\paragraph{Impact of Hyper-parameters.} Figure~\ref {fig:exp/sensitivity} evaluates the hyperparameters of AgenticGEO on GEO-Bench.
Multi-turn rewriting works best at $3$ turns (overall $25.48$), while adding more turns brings only small gains, indicating that a short planning is enough in practice. Archive size works best with a medium archive ($25$--$35$ strategies), peaking at $35$, whereas very small or very large archives perform worse, reflecting a trade-off between exploration and exploitation.

\begin{figure}[t]
  \centering

  \begin{minipage}{\columnwidth}
    \centering
    \captionof{table}{Offline critic ranking quality.}
    \label{tab:critic}
    \input{Tables/critic}
  \end{minipage}

  \vspace{6pt}

  \begin{minipage}{\columnwidth}
    \centering
    \includegraphics[width=0.65\columnwidth]{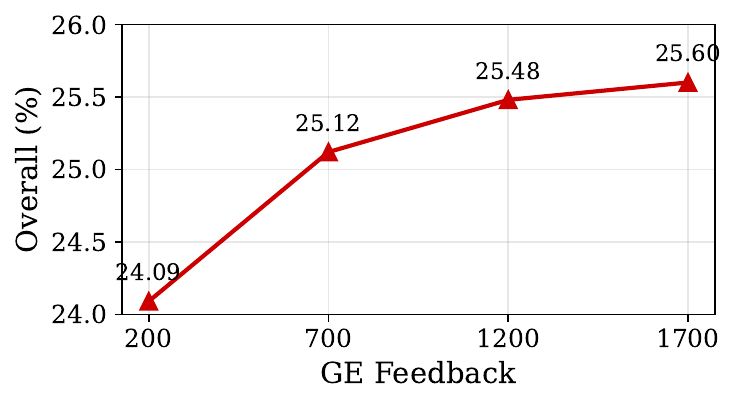}
    \captionof{figure}{Effect of the amount of GE feedback on overall performance when evolving. }
    \label{fig:exp/ge_feedback}
  \end{minipage}
\end{figure}

\begin{figure}[t]
  \centering
  \includegraphics[width=0.9\linewidth]{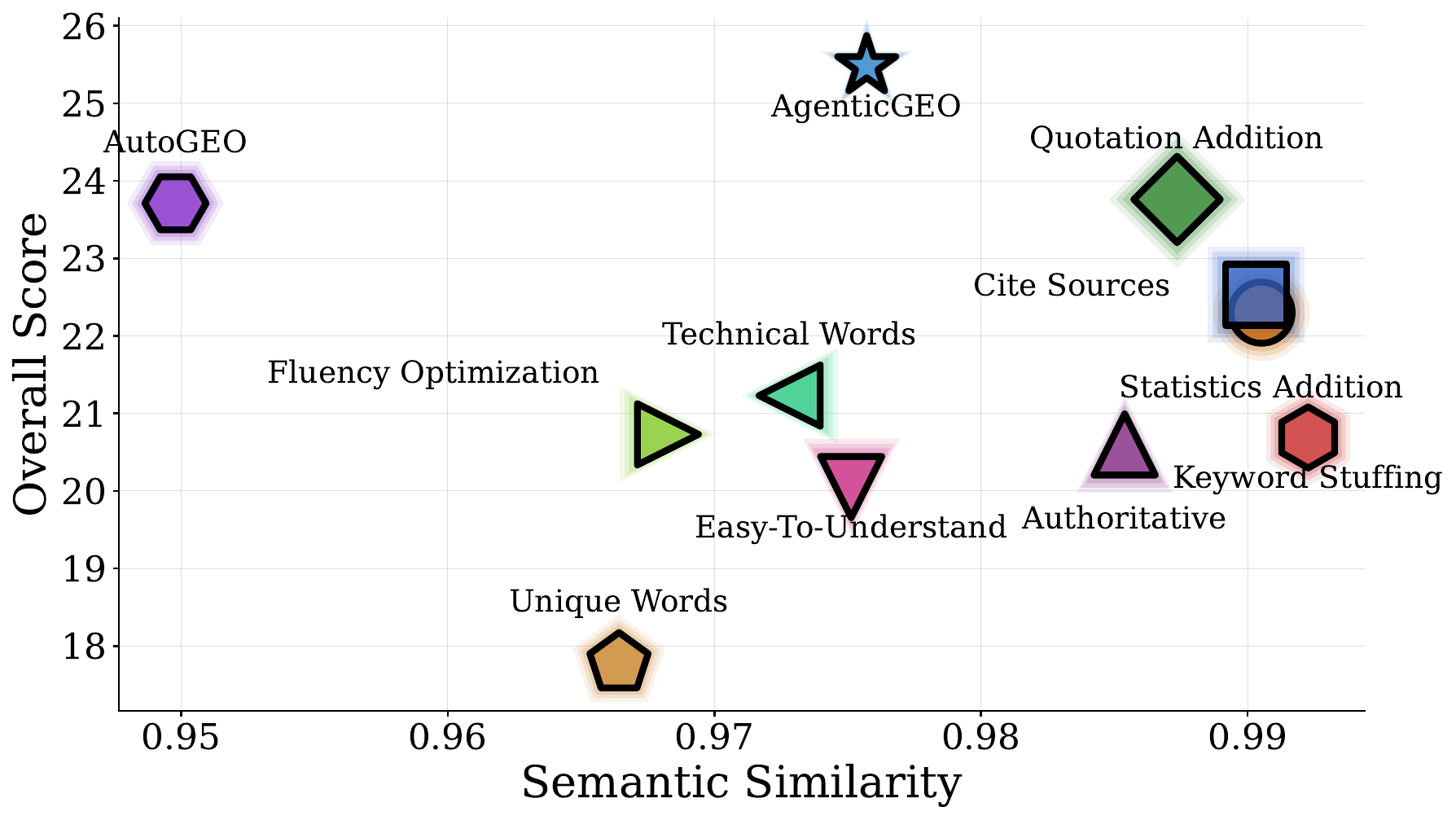}
  \caption{Semantic consistency and optimization effectiveness. Semantic Similarity is measured by BERTScore-F1~\cite{zhang2019bertscore} with \texttt{roberta-large}~\cite{liu2019roberta}, computed between the original content and the rewritten version.} 
  \label{fig:semantic_consistency}
\end{figure}

\paragraph{Impact of Offline Critic Alignment.}
We evaluate the efficacy of offline pre-training as a warm-start for online co-evolutionary learning. The critic is prompted to predict how the downstream engine would rank the seed strategies, comparing to the ground-truth with NDCG@K~\cite{jarvelin2002cumulated} metrics. Table~\ref{tab:critic} shows that the offline-pretrained critic achieves consistently high NDCG across all datasets, including two unseen domains. On GEO-Bench, the critic closely matches the engine-derived preference ordering, with an NDCG@5 of approximately $95\%$. While performance degrades on unseen domains, the critic still preserves strong top-rank fidelity, suggesting that it captures transferable signals rather than overfitting to the training distribution. These findings validate that the offline aligned critic model can serve as a reliable surrogate evaluator. 

\paragraph{The role of critic as the  surrogate of GE at online evolution}
As Figure~\ref{fig:exp/ge_feedback} shows, by using the critic as a low-cost surrogate of the GE environment, we can substantially reduce expensive GE interactions without sacrificing much performance. 
With only $700$ GE feedback, our method reaches an overall score of $25.12$, preserving $98.1\%$ of the best performance ($25.60$) while using only $41.2\%$ of the GE supervision, demonstrating sample-efficient online evolution under a limited feedback budget.

\subsection{Semantic Consistency Evaluation (RQ4)}

Figure~\ref{fig:semantic_consistency} compares the trade-off between semantic similarity and gains in overall scores. 
Most heuristic baselines preserve semantics well but yield limited improvements, suggesting that small edits alone are often insufficient to meaningfully increase a document's influence on visibility and attribution in the synthesized answers.  
AutoGEO achieves stronger overall performance, yet its semantic similarity is noticeably lower, indicating that its gains may come at the cost of larger content drift and information loss. 
In contrast, AgenticGEO attains the best overall score while maintaining relatively high semantic similarity, demonstrating that it can strengthen a document's impact on the generated answers without information loss. 
This indicates that AgenticGEO does not rely on aggressive rewriting. It leverages content-aware strategy selection and moderate, targeted edits that enhance salience and evidence presentation while largely preserving the original meaning.




%% file: Tables/exp1.tex
\begin{table}[t]
\centering

\renewcommand{\arraystretch}{1.15}
\caption{Overall Performance on the in-domain setting. Average results on $5$ independent runs are reported. $*$ indicates the statistically significant improvements over the best baseline, with $p$-value smaller than $0.001$.}
\label{tab:exp/geo-bench}
\vspace{-2ex}
\small
\resizebox{\columnwidth}{!}{
\begin{tabular}{l
                c@{\hskip 3pt}c@{\hskip 3pt}c
                @{\hskip 10pt}
                c@{\hskip 3pt}c@{\hskip 3pt}c}
\toprule
\multirow[b]{3}{*}{\textbf{Methods}} & 
\multicolumn{6}{c}{\textbf{GEO-Bench}} \\
\cmidrule(lr){2-7}
& \multicolumn{3}{c}{\textbf{Qwen2.5-32B-Instruct}} 
& \multicolumn{3}{c}{\textbf{Llama3.3-70B-Instruct}} \\
\cmidrule(lr){2-4}\cmidrule(lr){5-7}
                                  & word & pos & overall & word & pos & overall \\ \hline
No optimization                   & 20.05 & 20.26 & 20.21  & 19.19 & 19.33 & 19.20 \\
Keyword Stuffing                  & 20.73 & 20.86 & 20.69  & 19.99 & 20.16 & 20.02 \\
Unique Words                      & 17.59 & 17.94 & 17.78  & 16.78 & 16.66 & 16.56 \\
Easy-To-Understand                & 20.10 & 20.19 & 20.05  & 18.72 & 18.93 & 18.85 \\
Authoritative                     & 20.41 & 20.93 & 20.60  & 19.41 & 19.48 & 19.47 \\
Technical Words                   & 21.22 & 20.97 & 21.23  & 19.55 & 19.59 & 19.50 \\
Fluency Optimization              & 20.66 & 20.85 & 20.73  & 19.31 & 19.58 & 19.47 \\
Cite Sources                      & 22.64 & 22.91 & 22.53  & 21.95 & 22.11 & 21.98 \\
Quotation Addition                & 23.96 & 24.18 & 23.76  & 21.74 & 21.77 & 21.57 \\
Statistics Addition               & 22.34 & 22.86 & 22.30  & 21.07 & 21.23 & 21.05 \\
\midrule
AutoGEO                           & 23.51 & 23.70 & 23.71  & \underline{22.77} & \underline{22.65} & \underline{22.78} \\
Cite Sources-SFT                  & 23.02 & 23.30 & 22.91  & 22.26 & 22.43 & 22.21 \\
Quotation Addition-SFT           & \underline{24.10} & \underline{24.28} & \underline{23.92}  & 22.31 & 22.45 & 22.20 \\
Statistics Addition-SFT          & 23.05 & 23.47 & 23.02  & 21.79 & 21.90 & 21.75 \\
\midrule
\textbf{\model(ours)*}                 & \textbf{25.42} &  \textbf{25.85} &  \textbf{25.48}  &  \textbf{24.38} &  \textbf{24.59} &  \textbf{24.52} \\ 
\rowcolor{mygray} \textbf{Gains ($\%$)}
& \textit{26.78}\hspace{0.25em} & \textit{27.59}\hspace{0.25em} & \textit{26.08}\hspace{0.25em}
& \textit{27.05}\hspace{0.25em} & \textit{27.21}\hspace{0.25em} & \textit{27.71} \\
\bottomrule
\end{tabular}}
\end{table}

%% file: Tables/exp2.tex
\begin{table*}[h]
\centering
\renewcommand{\arraystretch}{1.08}
\caption{Overall Performance on the cross-domain setting. Average results on $5$ independent runs are reported. $*$ indicates the statistically significant improvements over the best baseline, with $p$-value smaller than $0.001$.}
\label{tab:exp/cross-domain}
\resizebox{0.7\textwidth}{!}{
\begin{tabular}{l
  r@{\hspace{4pt}}r@{\hspace{4pt}}r  @{\hspace{12pt}}
  r@{\hspace{4pt}}r@{\hspace{4pt}}r  @{\hspace{12pt}}
  r@{\hspace{4pt}}r@{\hspace{4pt}}r  @{\hspace{12pt}}
  r@{\hspace{4pt}}r@{\hspace{4pt}}r
}
\toprule

& \multicolumn{6}{c}{\textbf{Qwen2.5-32B-Instruct}}
& \multicolumn{6}{c}{\textbf{Llama3.3-70B-Instruct}} \\
\cmidrule(lr){2-7}\cmidrule(lr){8-13}

\textbf{Methods}
& \multicolumn{3}{c}{\textbf{MS MARCO}}
& \multicolumn{3}{c}{\textbf{E-Commerce}}
& \multicolumn{3}{c}{\textbf{MS MARCO}}
& \multicolumn{3}{c}{\textbf{E-Commerce}} \\
\cmidrule(lr){2-4}\cmidrule(lr){5-7}\cmidrule(lr){8-10}\cmidrule(lr){11-13}

& \multicolumn{1}{c}{word} & \multicolumn{1}{c}{pos} & \multicolumn{1}{c}{overall}
& \multicolumn{1}{c}{word} & \multicolumn{1}{c}{pos} & \multicolumn{1}{c}{overall}
& \multicolumn{1}{c}{word} & \multicolumn{1}{c}{pos} & \multicolumn{1}{c}{overall}
& \multicolumn{1}{c}{word} & \multicolumn{1}{c}{pos} & \multicolumn{1}{c}{overall} \\
\midrule

No optimization        & 19.99 & 20.15 & 20.05 & 18.30 & 18.09 & 18.01 & 19.45 & 19.64 & 19.67 & 19.70 & 19.45 & 19.68 \\
Keyword Stuffing       & 23.26 & 22.63 & 22.75 & 18.94 & 18.52 & 18.65 & 22.02 & 21.77 & 21.96 & 20.24 & 19.96 & 20.16 \\
Unique Words           & 17.95 & 18.24 & 18.23 & 18.12 & 18.09 & 18.01 & 18.94 & 18.88 & 18.76 & 19.16 & 19.04 & 19.15 \\
Easy-To-Understand     & 20.48 & 20.46 & 20.46 & 20.46 & 20.45 & 20.29 & 19.75 & 19.84 & 19.84 & 20.28 & 19.91 & 20.18 \\
Authoritative          & 21.29 & 21.08 & 21.07 & 19.84 & 19.32 & 19.64 & 20.23 & 20.21 & 20.10 & 20.06 & 19.82 & 20.06 \\
Technical Words        & 22.20 & 22.15 & 22.27 & 20.58 & 20.68 & 20.34 & 21.59 & 21.51 & 21.58 & 20.53 & 20.22 & 20.48 \\
Fluency Optimization   & 19.59 & 19.20 & 19.41 & 20.25 & 20.06 & 20.11 & 20.99 & 20.95 & 20.97 & 19.75 & 19.54 & 19.68 \\
Cite Sources           & 27.65 & 26.47 & 26.54 & 21.28 & 21.06 & 21.54 & 25.36 & 24.71 & 24.97 & 21.69 & 21.31 & 21.52 \\
Quotation Addition     & 29.70 & 28.70 & 28.43 & 21.75 & 21.91 & 21.54 & 26.59 & 25.81 & 25.66 & 20.85 & 20.49 & 20.69 \\
Statistics Addition    & 25.79 & 24.89 & 24.91 & 20.64 & 20.15 & 20.43 & 23.69 & 23.17 & 23.33 & 20.29 & 20.07 & 20.28 \\
\midrule
AutoGEO                & \underline{31.79} & \underline{31.14} & \underline{30.67} & 21.54 & 19.75 & 21.18 & \underline{30.27} & \underline{29.11} & \underline{30.04} & 21.45 & 20.13 & 21.50 \\
Cite Sources-SFT       & 30.24 & 29.55 & 29.36 & 21.96 & 21.88 & 21.67 & 28.63 & 28.07 & 28.15 & 21.70 & 21.43 & 21.65 \\
Quotation Addition-SFT & 31.16 & 30.30 & 30.08 & \underline{22.14} & \underline{21.96} & \underline{21.83} & 29.57 & 28.91 & 28.96 & \underline{21.91} & \underline{21.66} & \underline{21.70} \\
Statistics Addition-SFT& 29.21 & 28.75 & 28.34 & 21.83 & 21.49 & 21.60 & 27.74 & 27.31 & 27.52 & 21.47 & 21.18 & 21.30 \\
\midrule
\textbf{AgenticGEO(ours)*} & \textbf{34.96} & \textbf{34.25} & \textbf{34.10} & \textbf{26.79} & \textbf{26.57} & \textbf{26.58}
                         & \textbf{33.63} & \textbf{33.82} & \textbf{33.50} & \textbf{26.38} & \textbf{26.63} & \textbf{26.88} \\
\rowcolor{mygray} \textbf{Gains ($\%$)}
& \textit{74.89}\hspace{0.15em} & \textit{69.98}\hspace{0.15em} & \textit{70.07}
& \textit{46.39}\hspace{0.15em} & \textit{46.88}\hspace{0.15em} & \textit{47.58}
& \textit{72.90}\hspace{0.15em} & \textit{72.20}\hspace{0.15em} & \textit{70.31}
& \textit{33.91}\hspace{0.15em} & \textit{36.92}\hspace{0.15em} & \textit{36.59} \\

\bottomrule
\end{tabular}}
\end{table*}

%% file: Tables/critic.tex

\small
\begin{tabular}{lccc}
  \toprule
  \textbf{Benchmark} & \textbf{NDCG@1} & \textbf{NDCG@3} & \textbf{NDCG@5} \\
  \midrule
  \textbf{GEO-Bench} & 84.01 & 93.89 & 94.98 \\
  \textbf{Ms-Marco}  & 77.73 & 81.39 & 82.82 \\
  \textbf{E-Commerce} & 68.47 & 73.77 & 78.46 \\
  \bottomrule
\end{tabular}

%% file: 6conclusion.tex
\vspace{-0.5ex} \section{Conclusion} We study Generative Engine Optimization (GEO) for black-box engines, shifting the objective from rank to visibility in synthesized outputs. We show that static heuristics lack adaptability under content heterogeneity and changing GE behaviors. To address high interaction costs, we introduce a lightweight, calibrated critic as a reliable proxy. Built on this, AgenticGEO enables content-adaptive, self-evolving optimization by co-evolving a diverse strategy archive with the critic. Across representative settings, AgenticGEO delivers consistent improvements and strong cross-domain transfer. Our study points to a sustainable direction for web ecosystem governance that rewards quality and diversity, fostering a mutually beneficial development for creators and engines.

%% file: appendix_A.tex
\section{Supplementary Information}
\subsection{Methodological Details}
\label{sec:method_details}

\tcbset{
  promptbox/.style={
    colback=black!2,
    colframe=black!25,
    boxrule=0.5pt,
    arc=2pt,
    left=6pt,right=6pt,top=4pt,bottom=4pt,
    breakable,
    listing only,
    listing options={
     basicstyle=\ttfamily\footnotesize,
     breaklines=true,
     breakatwhitespace=false, 
     columns=fullflexible,
     keepspaces=true,
     showstringspaces=false
    }
  }
}

\tcbset{
  promptboxtext/.style={
    colback=black!2,
    colframe=black!25,
    boxrule=0.5pt,
    arc=2pt,
    left=6pt,right=6pt,top=4pt,bottom=4pt,
    breakable,
    enhanced,
    fontupper=\ttfamily\footnotesize,
    before upper=\raggedright,
    boxsep=0pt
  }
}
\subsubsection{Explanation of Figure\ref{fig:obs}}
\label{app:fig1}

For each instance, we run the nine rewriting strategies and obtain their average overall scores $\{r_i\}_{i=1}^{9}$, with the best score $r^{\star}=\max_i r_i$.
We quantify how many strategies remain competitive relative to the best. A strategy is considered near-optimal if it achieves at least $55\%$ of $r^{\star}$ (equivalently, its gap to $r^{\star}$ is no more than $45\%$ of $r^{\star}$).
Sensitivity is defined as the complement of this near-optimal fraction:
\[
\text{Sensitivity}
= 1-\frac{1}{9}\sum_{i=1}^{9}\mathbb{I}\!\left[r_i \ge 0.55\, r^{\star}\right]\in[0,1].
\]
Higher values indicate that only a few strategies are competitive (high sensitivity), whereas lower values suggest many strategies perform similarly well (low sensitivity).

With normalized sensitivity on the x-axis and maximum gain on the y-axis, we split instances into four regions.
\emph{(i) Robustly Optimizable}: many strategies achieve similarly high gains.
\emph{(ii) Strategy-Dependent}: high gains exist, but only a few strategies work well.
\emph{(iii) Optimization-Resistant}: strategies behave similarly and gains remain small.
\emph{(iv) Low-Yield \& Volatile}: outcomes vary widely, yet the best gain is still small.

The figure gives us two insights. First, GEO is instance-dependent, so a fixed strategy pool is unreliable. Second, \emph{Strategy-Dependent} region motivates content-conditioned strategy selection.

\subsubsection{Cited answer output}
The engine output is a cited answer
$\mathcal{A}=\{(y_i,\mathcal{C}_i)\}_{i=0}^{L-1}$,
where $L$ is the number of generated sentences,
$y_i$ is the $i$-th sentence,
and $\mathcal{C}_i\subseteq\{1,\dots,n\}$ denotes the indices of candidate documents cited in $y_i$.
Let $j^\star$ denote the index of the optimized content $d'$ within the candidate set.
To standardize $\mathcal{E}$'s cited-answer generation, we use the following prompt template.

\begin{tcolorbox}[promptbox, title={Engine Answer Synthesis Prompt Template}]
Write an accurate and concise answer for the given user question.
using only the provided summarized web search results.
The answer should be correct, high-quality, and written by an expert
using an unbiased and journalistic tone.
The user's language of choice, such as English, Français, Español, or Deutsch
should be used.
The answer should be informative, interesting, and engaging.
The answer's logic and reasoning should be rigorous and defensible.
Every sentence in the answer should be immediately followed by an in-line
citation to the search result(s).
The cited search result(s) should fully support all the information in the
sentence.
Search results need to be cited using [index].
When citing several search results, use [1][2][3] format rather than [1, 2, 3].
You can use multiple search results to respond comprehensively while avoiding
irrelevant search results.
\end{tcolorbox}

\subsubsection{Impression metrics}
\label{app:metrics}

Following GEO-Bench~\cite{aggarwal2024geo}, we quantify the visibility of each candidate document $j \in \{1, \dots, n\}$ within the generated response $\mathcal{A} = \{y_0, \dots, y_{L-1}\}$ by aggregating its attributed contributions. 
Here, $y_i$ denotes the $i$-th sentence in $\mathcal{A}$, and $\mathcal{C}_i$ denotes the set of citation indices associated with $y_i$.
When a sentence cites multiple candidates, its contribution is uniformly distributed by a factor of $1/|\mathcal{C}_i|$.

Let $\mathrm{wc}(y_i)$ be the word count of sentence $y_i$. To reflect user attention decay, we define a position weight $w(i)$ for the $i$-th sentence:
\begin{equation}
w(i)=
\begin{cases}
\exp\!\left(-\frac{i}{L-1}\right), & L>1,\\
1, & L=1.
\end{cases}
\end{equation}

We compute three impression scores: \textsc{word} (attributed word count), \textsc{pos} (citation order with position weights), and \textsc{overall} (a combination of word count and position decay):
\begin{align}
\mathrm{Score}^{\textsc{word}}_j(q,s) &=
\sum_{i=0}^{L-1}\mathbb{I}[j\in\mathcal{C}_i]\cdot
\frac{\mathrm{wc}(y_i)}{|\mathcal{C}_i|}, \\
\mathrm{Score}^{\textsc{pos}}_j(q,s) &=
\sum_{i=0}^{L-1}\mathbb{I}[j\in\mathcal{C}_i]\cdot
\frac{w(i)}{|\mathcal{C}_i|}, \\
\mathrm{Score}^{\textsc{overall}}_j(q,s) &=
\sum_{i=0}^{L-1}\mathbb{I}[j\in\mathcal{C}_i]\cdot
\frac{\mathrm{wc}(y_i)\cdot w(i)}{|\mathcal{C}_i|}.
\end{align}

The goal of GEO is to maximize the impression of the optimized content $\tilde{d}$ in the generative engine output $\mathcal{A}$, as quantified by the above metrics.

\subsubsection{Seed Strategies}
\label{app:seed}
We initialize the critic's offline preference alignment with 9 seed rewriting strategies.
Each seed prompt is a template applied to the source summary (placeholder \texttt{\{summary\}}) to produce candidate rewrites, which are then used to construct offline preference data for warm-starting the critic.






\begin{tcolorbox}[promptbox, title={Keyword Stuffing}]
\textbf{Task:} 
Improve the source by inserting up to 10 NEW, relevant SEO keywords that are NOT already present in the text.

\textbf{Constraints:}
\begin{itemize}[label={--}, leftmargin=15pt, topsep=2pt, itemsep=0pt, parsep=0pt]
    \item Do not change, add, or remove any core information.
    \item Keep the original structure (paragraphing, bullet points, line breaks).
    \item Insert keywords naturally inline (no keyword list at the end).
\end{itemize}

\textbf{Source:} \texttt{summary}

\textbf{Output:} The updated source text only.
\end{tcolorbox}

\begin{tcolorbox}[promptbox, title={Unique Words}]
\textbf{Task:} Revise the source by using more unique and precise vocabulary.

\textbf{Constraints:}
\begin{itemize}[label={--}, leftmargin=15pt, topsep=2pt, itemsep=0pt, parsep=0pt]
    \item Preserve the original meaning and all core information.
    \item Do not add new claims or remove any content.
    \item Keep the length and structure roughly the same.
\end{itemize}

\textbf{Source:} \texttt{summary}

\textbf{Output:} The revised source text only.
\end{tcolorbox}

\begin{tcolorbox}[promptbox, title={Easy-To-Understand}]
\textbf{Task:} Rewrite the source in simple, easy-to-understand language.

\textbf{Constraints:}
\begin{itemize}[label={--}, leftmargin=15pt, topsep=2pt, itemsep=0pt, parsep=0pt]
    \item Do not omit, add, or alter any core information.
    \item Keep the original structure and roughly the same length.
    \item Only rephrase sentences for clarity and readability.
\end{itemize}

\textbf{Source:} \texttt{summary}

\textbf{Output:} The simplified source text only.
\end{tcolorbox}

\begin{tcolorbox}[promptbox, title={Authoritative}]
\textbf{Task:} Make the source sound confident, authoritative, and expert.

\textbf{Constraints:}
\begin{itemize}[label={--}, leftmargin=15pt, topsep=2pt, itemsep=0pt, parsep=0pt]
    \item Do not add new facts or remove any information.
    \item Keep the original structure (formatting, bullets, spacing).
    \item Strengthen tone via wording choices, not by exaggerating or making unverifiable claims.
\end{itemize}

\textbf{Source:} \texttt{summary}

\textbf{Output:} The revised source text only.
\end{tcolorbox}

\begin{tcolorbox}[promptbox, title={Technical Words}]
\textbf{Task:} Rewrite the source in a more technical style using domain-appropriate terminology.

\textbf{Constraints:}
\begin{itemize}[label={--}, leftmargin=15pt, topsep=2pt, itemsep=0pt, parsep=0pt]
    \item Preserve all core information; do not introduce new claims.
    \item Keep the structure and length roughly unchanged.
    \item Rephrase sentences to sound more technical and precise.
\end{itemize}

\textbf{Source:} \texttt{summary}

\textbf{Output:} The revised source text only.
\end{tcolorbox}

\begin{tcolorbox}[promptbox, title={Fluency Optimization}]
\textbf{Task:} Rewrite the source to improve fluency and coherence.

\textbf{Constraints:}
\begin{itemize}[label={--}, leftmargin=15pt, topsep=2pt, itemsep=0pt, parsep=0pt]
    \item Do not alter the core content.
    \item Improve sentence transitions and readability.
    \item Keep the structure and length roughly the same.
\end{itemize}

\textbf{Source:} \texttt{summary}

\textbf{Output:} The rewritten source text only.
\end{tcolorbox}

\begin{tcolorbox}[promptbox, title={Cite Sources}]
\textbf{Task:} Strengthen credibility by adding a small number of natural-language citations to credible sources (e.g., industry reports, standards, official docs).

\textbf{Constraints:}
\begin{itemize}[label={--}, leftmargin=15pt, topsep=2pt, itemsep=0pt, parsep=0pt]
    \item Citations must be plausible and verifiable; do not fabricate sources.
    \item Do not change the core information or add new claims.
    \item Keep structure and length roughly the same (about 5--6 citations total).
\end{itemize}

\textbf{Source:} \texttt{summary}

\textbf{Output:} The revised source text only.
\end{tcolorbox}

\begin{tcolorbox}[promptbox, title={Quotation Addition}]
\textbf{Task:} Increase perceived authority by adding a few short, relevant quotations from reputable entities (e.g., well-known organizations or experts).

\textbf{Constraints:}
\begin{itemize}[label={--}, leftmargin=15pt, topsep=2pt, itemsep=0pt, parsep=0pt]
    \item Quotes must be accurate and attributable; do not invent quotes.
    \item Do not change core content; keep structure and length similar.
    \item Integrate quotes inline without adding long new paragraphs.
\end{itemize}

\textbf{Source:} \texttt{summary}

\textbf{Output:} The revised source text only.
\end{tcolorbox}

\begin{tcolorbox}[promptbox, title={Statistics Addition}]
\textbf{Task:} Add a few concise, relevant statistics or numerical facts to improve concreteness.

\textbf{Constraints:}
\begin{itemize}[label={--}, leftmargin=15pt, topsep=2pt, itemsep=0pt, parsep=0pt]
    \item Statistics must be verifiable; do not invent numbers.
    \item Do not modify core content beyond inserting stats inline.
    \item Keep the original structure and stop at the end of the original source.
\end{itemize}

\textbf{Source:} \texttt{summary}

\textbf{Output:} The revised source text only.
\end{tcolorbox}

\subsubsection{Genotype Details}
\label{subsec:app_genotype}
We formalize the evolving strategy as a structured genotype $g = \langle g^I, g^C, g^R, g^F, g^T \rangle$. To interface efficiently with the Critic and the Generative Engine, we implement two deterministic rendering functions $R_\text{crit}$ and $R_\text{eng}$ :

\noindent \textbf{1. Compact Summary for Critic ($R_{\text{crit}}$).} To minimize token consumption while retaining discriminative features, $R_{\text{crit}}(g)$ maps the genotype to a concatenated string of active categorical values. Formally, let $\mathcal{K}_{\text{active}} \subset g$ be the set of non-empty discrete fields (e.g., tone labels, format types). The rendering is defined as:
\begin{equation}
R_{\text{crit}}(g) = \bigoplus_{k \in \mathcal{K}_{\text{active}}} \left( \text{Name}(k) || \text{":"} || \text{Val}(k) \right),
\end{equation}
where $\oplus$ denotes string concatenation with delimiters. For example, a strategy might be rendered as ``\texttt{Tone:Assertive|Format:List|
Constraint:Anti-Hallucination}''.

\noindent \textbf{2. Full Prompt for Engine ($R_{\text{eng}}$).} $R_{\text{eng}}(g)$ acts as a template-filling function that constructs the executable meta-prompt. It wraps the raw text of each gene component into specific sections:
\begin{equation}
R_{\text{eng}}(g) = \mathcal{T}_{\text{sys}} \oplus g^I \oplus \mathcal{T}_{\text{cons}}(g^C) \oplus \mathcal{T}_{\text{reason}}(g^R) \oplus \mathcal{T}_{\text{fmt}}(g^F) \oplus \mathcal{T}_{\text{tone}}(g^T),
\end{equation}
where $\mathcal{T}$ represents fixed instructional templates (e.g., ``\textit{Adhere to the following constraints: ...}''). This full prompt is then combined with the query $q$ and content $d$ to form the final input for the rewriting model.

\subsubsection{MAP-Elites Descriptors and Archive Gates}
\label{subsec:app_gates}
The strategy space is discretized into behavioral cells via a descriptor function $\psi: \mathcal{G} \to \mathbb{Z}^D$. Based on our design, $\psi(g)$ maps a genotype to a tuple of 12 discrete dimensions :
\begin{itemize}[leftmargin = 8pt]
    \item \textbf{Core Types:} \texttt{strategy\_type}, \texttt{output\_schema}.
    \item \textbf{Switches:} \texttt{has\_self\_check}, \texttt{has\_reasoning}, \texttt{has\_conflict\_res}, \texttt{use\_code\_block}, \texttt{has\_prelude}, \texttt{has\_post\_check}.
    \item \textbf{Buckets:} \texttt{tone\_bucket}, \texttt{constraint\_strength}, \texttt{length\_policy}, \texttt{reasoning\_steps\_bucket}.
\end{itemize}
A candidate strategy $s$ (with genotype $g$) is mapped to a cell index $c = \psi(g)$. It is admitted only if it passes two gates:

\noindent \textbf{1. Novelty Gate (De-duplication).}
We de-duplicate candidates using character-level $n$-gram Jaccard similarity computed on the rendered strategy summaries. The similarity between a candidate $s$ and an existing elite $e$ is:
\begin{equation}
\operatorname{Sim}(s, e)
= \frac{|\text{$n$-grams}(s)\cap \text{$n$-grams}(e)|}{|\text{$n$-grams}(s)\cup \text{$n$-grams}(e)|}.
\end{equation}
The candidate is rejected if it is too similar to any strategy already stored in the target cell:
\[
\max_{e}\operatorname{Sim}(s,e) > 0.9,
\]
which prevents near-duplicates.

\noindent \textbf{2. Value Gate (Performance).}
If the target cell is not full ($<K_c$ strategies), $s$ is admitted. Otherwise, it must beat the current worst strategy in that cell.

\subsubsection{PND Score Formulation and Pruning}
\label{subsec:app_pnd}
We maintain the archive using a \textbf{PND score} that balances effectiveness and exploration:
\begin{equation}
S_{\mathrm{PND}}(s)= r(s) + \lambda_{\mathrm{pnd}}\big(\operatorname{Nov}(s)+\operatorname{Div}(s)\big),
\label{eq:pnd}
\end{equation}
where $r(s)$ is the impression score gain evaluated by the critic or GE.

\noindent \textbf{Novelty ($\operatorname{Nov}$).}
$\operatorname{Nov}(s)$ encourages population coverage by rewarding strategies that are structurally dissimilar to those already stored in the current archive $\mathcal{M}$ (using similarity metric as Eq.~(21)).

\noindent \textbf{Diversity ($\operatorname{Div}$).}
$\operatorname{Div}(s)$ promotes diverse evolutionary trajectories by favoring strategies with richer lineage history (e.g., deeper generations), more varied mutation operators, and less degenerate genotypes (more fields actively used).

\subsubsection{Evolver Action Space and Prompting}
\label{subsec:app_evolver}

The Evolver $\pi_\psi$ functions as a meta-optimizer that proposes improvements to existing strategies. It takes as input an instance $x=(q,d)$, a primary parent $g_A$, an optional secondary parent $g_B$ (for crossover), and an operator catalog $\Omega$.

\noindent \textbf{Operator Catalog ($\Omega$).} To ensure diverse and controllable evolution, $\Omega$ consists of two categories of symbolic operators. 

\noindent \textbf{Mutation Operators} apply field-level perturbations targeting specific dimensions, such as \texttt{mut\_C\_strengthen} (adding constraints), \texttt{mut\_T\_toggle\_tone} (switching styles), and \texttt{mut\_F\_schema\_swap} (changing output format). 

\noindent \textbf{Crossover Operators} synthesize features from two parents, including \texttt{cx\_swap\_gene} (exchanging gene blocks) and \texttt{cx\_conflict\_sy-
nthesis} (resolving conflicts between Parent A and B).

\noindent \textbf{Action Proposal.} The Evolver outputs $M$ candidate actions. Each action is a strict JSON object $a=\{\texttt{operator\_id}, \texttt{child\_genotype}\}$, where \texttt{child\_genotype} is the full structure resulting from applying the selected operator.

\begin{tcolorbox}[promptbox, title={Evolver Action Proposal Prompt Template}]
System:
You are a prompt evolution agent for GEO. You must evolve a parent strategy (or combine two parents) into a better STRUCTURED GENOTYPE JSON (I/C/R/F/T).

1) Choose an operator\_id from the provided catalog.
2) Produce a child\_genotype JSON that results from applying that operator.

Important constraints:
- The output MUST be valid JSON (one object per line).
- The child genotype MUST preserve the I/C/R/F/T structure.
- If choosing a Crossover operator (starts with "cx\_"): You MUST conceptually combine Parent A and Parent B.
- If Parent B is NOT provided: Do NOT choose any "cx\_*" operator.
- Prefer DIVERSITY: Avoid repeating the same operator across candidates.

User:
\#\# Query
\{query\}

\#\# Document Summary
\{content\_summary\}

\#\# Parent Genotype A (JSON)
\{parent\_genotype\_json\}

\#\# Parent Genotype B (JSON) [Optional]
\{parent\_b\_genotype\_json\}

\#\# Operator Catalog
\{operator\_catalog\}

\#\# Task
Generate \{num\_candidates\} candidates. Output exactly \{num\_candidates\} JSON lines.
\end{tcolorbox}

\subsection{Experiment Details}
\label{app:exp}

\subsubsection{Dataset.}
\label{app:dataset}
To comprehensively evaluate AgenticGEO, we conduct experiments across three datasets characterized by distinct content distributions and optimization goals:

\begin{itemize}[leftmargin=*]
    \item \textbf{GEO-Bench (In-Domain)}: 
    This serves as our primary dataset for training the evolver and critic, as well as for in-domain evaluation. Constructed from real-world Google Search results, it covers a wide spectrum of domains (e.g., Science, History, Health) with varying query difficulties. The content primarily consists of long-form articles, providing rich context for learning diverse optimization strategies.
    
    \item \textbf{MS MARCO (Out-of-Domain)}: 
    To assess zero-shot transferability, we employ the MS MARCO Passage Ranking dataset. Derived from Bing search logs, this dataset comprises real user queries paired with short, often unstructured text passages. Evaluating on this dataset tests whether our optimization policy can generalize to short-text scenarios and unseen query distributions without re-training.
    
    \item \textbf{E-commerce (Vertical Domain)}: 
    Representing a specific vertical application, this dataset is sourced from Amazon product descriptions and reviews. The optimization goal here shifts from informational retrieval to commercial visibility. This dataset challenges the agent to adapt to entity-centric content and the specific structural preferences of product-related queries.
\end{itemize}

\begin{table}[t]
\centering
\small
\setlength{\tabcolsep}{7pt}
\caption{Dataset statistics after preprocessing. Following the GEO-Bench protocol, each query is paired with $5$ documents.}
\label{tab:dataset_stats}
\begin{tabular}{lrrr}
\toprule
\textbf{Dataset} & \textbf{\#Queries} & \textbf{\#Docs} & \textbf{Avg. Content Tokens} \\
\midrule
GEO-Bench  & 1000 & 5{,}000 & 980.61 \\
MS-Marco   & 1000 & 5{,}000 & 91.91 \\
E-commerce & 416 & 2{,}180 & 1{,}459.69 \\
\bottomrule
\end{tabular}
\end{table}

Table~\ref{tab:dataset_stats} reports the key statistics of the three datasets.
Following the GEO-Bench protocol, we standardize the retrieval context by pairing each query with five documents, ensuring a consistent document budget across datasets for fair comparison.

\subsubsection{Baselines.}
\label{app:baseline}

\mbox{}

\noindent \textbf{1. Static Heuristics (GEO-Bench).}
We implement the nine official strategies from GEO-Bench~\cite{aggarwal2024geo}, covering lexical, stylistic, and evidence-based modifications:
\begin{itemize}[leftmargin = 8pt]
    \item \textbf{No Optimization:} The original, unmodified source content serves as the control group.
    \item \textbf{Keyword Stuffing:} Naively injects query keywords repeatedly to increase term frequency.
    \item \textbf{Unique Words:} Inserts rare vocabulary to artificially increase information entropy.
    \item \textbf{Easy-To-Understand:} Simplifies sentence structures to improve readability for general audiences.
    \item \textbf{Authoritative:} Adopts a confident and professional tone to mimic expert knowledge.
    \item \textbf{Technical Words:} Injects domain-specific jargon assuming engines prefer specialized vocabulary.
    \item \textbf{Fluency Optimization:} Polishes the text for grammatical correctness without adding information.
    \item \textbf{Cite Sources:} Injects plausible citations to external authorities to enhance credibility.
    \item \textbf{Quotation Addition:} Embeds direct quotes from relevant entities to support claims.
    \item \textbf{Statistics Addition:} Enriches the text with quantitative data points relevant to the query.
\end{itemize}

\noindent \textbf{2. State-of-the-Art.}
\begin{itemize}[leftmargin=8pt]
    \item \textbf{AutoGEO~\cite{autogeo}:} A representative automated framework that distills generative engine preferences from LLM-generated explanations into static rewriting rules. To reimplement the method, we use the GEO-Bench training dataset on the generative engine of Qwen2.5-32B-Instruct following the source code of AutoGEO.
\end{itemize}

\noindent \textbf{3. Supervised Fine-Tuning (SFT).}
To compare with learning-to-rewrite baselines, we fine-tune a rewriter on supervised pairs (where the target rewrites are selected based on overall score), targeting a single heuristic style, yielding controllable specialized rewriters:
\begin{itemize}[leftmargin=8pt]
    \item \textbf{Cite Sources-SFT:} Fine-tunes a rewriter to produce citation-enriched rewrites in the style of \textit{Cite Sources}.
    \item \textbf{Quotation Addition-SFT:} Fine-tunes a rewriter to add concise supporting quotations in the style of \textit{Quotation Addition}.
    \item \textbf{Statistics Addition-SFT:} Fine-tunes a rewriter to insert relevant numeric statements in the style of \textit{Statistics Addition}.
\end{itemize}

\subsubsection{Hyper Parameters}
\input{Tables/hyper}

Key hyperparameters are in Table~\ref{tab:hyperparams}.

%% file: Tables/hyper.tex
\begin{table}[t]
\centering
\footnotesize
\renewcommand{\arraystretch}{0.98}
\setlength{\tabcolsep}{3.2pt}
\caption{Key hyperparameters of AgenticGEO (values left blank).}
\label{tab:hyperparams}
\begin{tabular}{llp{0.52\linewidth}c}
\toprule
\textbf{Module} & \textbf{Param.} & \textbf{Description} & \textbf{Value} \\
\midrule

\multicolumn{4}{l}{\textbf{Offline Critic Alignment}} \\
Loss weight & $\lambda$ &
Weight in $L_{\text{total}}=L_{\text{pair}}+\lambda L_{\text{reg}}$ & 0.2 \\
Warm-up steps & $S_{\text{freeze}}$ &
Epochs with frozen backbone before unfreezing & 1 \\

\midrule
\multicolumn{4}{l}{\textbf{Online Co-Evolution (Selection \& Budget)}} \\
Iterations & $T$ & Total online evolution iterations & 100 \\
Exploit size & $K_{\text{top}}$ & Top-$K_{\text{top}}$ selected by critic & 4 \\
Explore size & $K_{\text{rand}}$ & Randomly sampled strategies per iteration &  4\\

\midrule
\multicolumn{4}{l}{\textbf{Evolver Learning (Sibling-Aware AWR)}} \\
Sibling coeff. & $\alpha_{\text{sib}}$ & Strength of sibling-aware baseline & 0.8\\
AWR temperature & $\beta$ & Temperature in $\exp(A/\beta)$ weighting & 1.0 \\

\midrule
\multicolumn{4}{l}{\textbf{Archive Maintenance (MAP-Elites \& Gates)}} \\
PND weight & $\lambda_{\text{pnd}}$ & Weight of novelty/diversity term in $S_{\text{PND}}$ & 0.3 \\

Cell capacity & $K_c$ & Max items stored per archive cell & 3 \\

\midrule
\multicolumn{4}{l}{\textbf{Implementation}} \\
LR (critic) & $\eta_c$ & Learning rate for critic fine-tuning & 0.001\\
LR (evolver) & $\eta_e$ & Learning rate for evolver fine-tuning & 0.0002\\
Batch size & $B$ & Batch size for fine-tuning &2 \\
LoRA rank & $r$ & LoRA rank & 16\\
LoRA scaling & $\alpha_{\text{lora}}$ & LoRA scaling factor & 32\\
LoRA dropout & $p_{\text{lora}}$ & LoRA dropout & 0.05\\
Epochs & $E$ & Fine-tuning epochs & 2\\

\bottomrule
\end{tabular}
\end{table}

%% file: appendix_B.tex
\vspace{-6pt}
\subsection{Theoretical Analysis}
\label{appendix:theorem}

To establish the bound, we make the standard assumptions for online convex optimization:
\begin{enumerate}
\item \textbf{Boundedness \& Lipschitz Continuity:} The true risk function $\mathcal{R}(\cdot)$ and the critic $\mathcal{C}_t(\cdot)$ are bounded in $[0, B]$ and are $L$-Lipschitz continuous with respect to the strategy parameters.
\item \textbf{Critic Generalization:} The critic is trained on an accumulating dataset $\mathcal{B}_t$.
\end{enumerate}

\begin{lemma}[Linear Growth of Replay Buffer]\label{lemma:buffer_growth}
Let $K = |\mathcal{S}_{\emph{select}}^{(t)}|$ denote the constant number of candidate strategies selected for ground-truth evaluation at iteration $t$. Assume that each candidate strategy has an independent success probability $p$. Only successful strategies will be added to the replay buffer $\mathcal{B}_T$. For any $0 < \delta < 1$, with probility at least $1-\delta$, the size of the replay buffer can be bounded by:
\begin{equation}
\big| \frac{|\mathcal{B}_T|}{pKT} -1 \big|  \le \sqrt{\frac{\log(\frac{2}{\delta})}{3pKT}}
\end{equation}
\begin{proof}
    Let $X_{t,i}$ be an indicator random variable representing the success of the $i$-th candidate strategy at iteration $t$, where $t \in \{1, \dots, T\}$ and $i \in \{1, \dots, K\}$. By assumption, $X_{t,i}$ are i.i.d. Bernoulli variables with parameter $p$. The replay buffer size can be written as
    \begin{equation}
    |\mathcal{B}_T| = \sum_{t=1}^{T} \sum_{i=1}^{K} X_{t,i}.
    \end{equation}
    Hence, the expected size of the buffer is:
    \begin{equation}
        \mu := \mathbb{E}[|\mathcal{B}T|] = \sum_{t=1}^{T} \sum_{i=1}^{K} \mathbb{E}[X_{t,i}] = T \cdot K \cdot p.
    \end{equation}
    By the standard Chernoff bound, for any $\epsilon \in (0, 1)$,
    \begin{equation}
        \mathbb{P}\big( \big| |\mathcal{B}_T| - \mu \big| \ge \epsilon \mu \big) \le 2 e^{-\frac{\epsilon^2 \mu}{3}}.
    \end{equation}
Setting the right-hand side equal to $\delta$ and solving for $\epsilon$ yields
    \begin{equation}
        \epsilon = \sqrt{\frac{3\log(\frac{2}{\delta})}{\mu}}
    \end{equation}
    Substituting $\mu = pKT$, we have
    \begin{equation}
        \mathbb{P}\big( \big| \frac{|\mathcal{B}_T|}{pKT} - 1 \big| \ge \sqrt{\frac{3\log(\frac{2}{\delta})}{\mu}} \big) \le \delta.
    \end{equation}   
\end{proof}
\end{lemma}

\begin{lemma}[Critic Generalization Bound]
\label{lemma:critic_bound}
Suppose the critic is learned from the replay buffer $\mathcal{B}_T$, and the hypothesis class $\mathcal{F}$ has bounded Rademacher complexity. Under the assumption of Lemma~\ref{lemma:buffer_growth}, for any $\delta \in (0, 1)$, with probability at least $1-\delta$,
\begin{equation}
    |\mathcal{C}_T(s) - \mathcal{R}(s)| \leq \frac{1}{\sqrt{T}} \left ( 2M_\mathcal{F}\sqrt{\frac{2}{pK}} + B\sqrt{\frac{\log(2/\delta)}{pK}}\right)
\end{equation}

\begin{proof}
The universal convergence bound states that with probability at least $1 - \frac{\delta}{2}$, the generalization error is bounded by
\begin{equation}
    |\mathcal{C}_T(s) - \mathcal{R}(s)| \leq 2\mathfrak{R}_{|\mathcal{B}_T|}(\mathcal{F}) + B\sqrt{\frac{\log(2/\delta)}{2|\mathcal{B}_T|}}.
\end{equation}
where $\mathfrak{R}_{|\mathcal{B}_T|} \leq \frac{M_\mathcal{F}}{|\mathcal{B}_T|}$ is the Rademacher complexity.
From Lemma~\ref{lemma:buffer_growth}, with probability at least $1 - \frac{\delta}{2}$, the buffer size 
\begin{equation}
    |\mathcal{B}_T| \geq (1 - \epsilon_T)pKT
\end{equation}
Substitute into the above generalization bound, we have with probability at least $1-\delta$
\begin{align} 
|\mathcal{C}_T(s) - \mathcal{R}(s)| & \leq \frac{1}{\sqrt{T}} \cdot \frac{1}{\sqrt{1 - \epsilon_T}} \left( 2M_\mathcal{F}\sqrt{\frac{1}{pK}} + B\sqrt{\frac{\log(2/\delta)}{2pK}} \right) 
\end{align}
For sufficiently large $T$ such that
\(
    T \geq \frac{8 \log(2/\delta)}{pK},
\)
we have $\epsilon_T \leq \frac{1}{2}$. Then we will have
\begin{align*}
    |\mathcal{C}_T(s) - \mathcal{R}(s)| & \leq \frac{1}{\sqrt{T}} \cdot \frac{1}{\sqrt{1 - \frac{1}{2}}} \left( 2M_\mathcal{F}\sqrt{\frac{1}{pK}} + B\sqrt{\frac{\log(2/\delta)}{2pK}} \right) \\
    & = \frac{1}{\sqrt{T}} \left( 2M_\mathcal{F}\sqrt{\frac{2}{pK}} + B\sqrt{\frac{\log(2/\delta)}{pK}} \right) \\
\end{align*}
This bound implies that the approximation error converges as $\mathcal{O}(\frac{1}{\sqrt{T}})$.
\end{proof}
\end{lemma}

\begin{lemma}[Evolver Regret Bound]\label{lemma:evolver_bound}
Let $\mathcal{C}_t : \mathcal{S} \to \mathbb{R}$ denote the risk predicted by the critic at iteration $t$. Assume that $\mathcal{C}_t$ is convex and $L$-Lipschitz continuous over the strategy space $\mathcal{S}$ with diameter $D$. Suppose the Evolver updates the strategy $s_t$ via a gradient-based update with step size $\eta$. Then, for any comparator strategy $s^\ast \in \mathcal{S}$, the cumulative regret with respect to the critic’s predictions satisfies
\begin{equation}
\sum_{t=1}^T \bigl( \mathcal{C}_t(s_t) - \mathcal{C}_t(s^\ast) \bigr)
\;\leq\;
DL\sqrt{T}
\;=\;
\mathcal{O}(\sqrt{T}),
\end{equation}
provided that the step size is chosen as $\eta = \tfrac{D}{L\sqrt{T}}$.
\begin{proof}
    The evolver updates the strategy with gradient-based approach:
    \begin{equation}
        s_{t+1} =  s_t - \eta g_t, \qquad \text{where}\ g_t = \nabla \mathcal{C}_t(s_t).
    \end{equation}
     For any $s^\ast$, we have 
     \begin{equation}
         \| s_{t+1} - s^\ast \|^2 \leq \| (s_t - \eta g_t) - s^\ast \|^2.
     \end{equation}
     We expand the RHS:
     \begin{equation}
        \| s_t - \eta g_t - s^\ast \|^2 = \| s_t - s^\ast \|^2 - 2\eta \langle g_t, s_t - s^\ast \rangle + \eta^2 \| g_t \|^2
     \end{equation}
     Rearranging this inequality:
     \begin{equation}
         2\eta \langle g_t, s_t - s^\ast \rangle \leq \| s_t - s^\ast \|^2 - \| s_{t+1} - s^\ast \|^2 + \eta^2 \| g_t \|^2
     \end{equation}
     Dividing by $2\eta$, we obtain
     \begin{equation}\label{eq:evolver-grad-decomposition}
         \langle g_t, s_t - s^\ast \rangle \leq \frac{1}{2\eta} \left( \| s_t - s^\ast \|^2 - \| s_{t+1} - s^\ast \|^2 \right) + \frac{\eta}{2} \| g_t \|^2
     \end{equation}
     From the first-order inequality, we have 
     \begin{equation}\label{eq:evolver-first-order-inequality}
        \mathcal{C}_t(s_t) - \mathcal{C}_t(s^\ast) \leq \langle g_t, s_t - s^\ast \rangle
    \end{equation}
    Combining and Eq~\eqref{eq:evolver-grad-decomposition} and Eq~\eqref{eq:evolver-first-order-inequality}, we sum from $t=1$ to $T$:
    \begin{align*}
       & \sum_{t=1}^T  \mathcal{C}_t(s_t) - \mathcal{C}_t(s^\ast) \leq \sum_{t=1}^T \langle g_t, s_t - s^\ast \rangle \\
        & \qquad  \leq \frac{1}{2\eta}\sum_{t=1}^T \left( \| s_t - s^\ast \|^2 - \| s_{t+1} - s^\ast \|^2 \right) + \frac{\eta}{2}\sum_{t=1}^T\|g_t\|^2
    \end{align*}
    For the first term, 
    \begin{equation}\sum_{t=1}^T \left( \| s_t - s^\ast \|^2 - \| s_{t+1} - s^\ast \|^2 \right) = \| s_1 - s^\ast \|^2 - \| s_{T+1} - s^\ast \|^2 \leq \| s_1 - s^\ast \|^2
    \end{equation}
    Note that the strategy space have a diameter of $D$ such that the distance $\| s_1 - s^\ast \|^2 \leq D^2$. Furthermore, since $\mathcal{C}_t$ is $L$-Lipschitz, the norm of the gradient is bounded by $\| g_t \| \leq L$. Thus,
    \begin{equation}
        \sum_{t=1}^T \langle g_t, s_t - s^\ast \rangle \leq \frac{D^2}{2\eta} + \frac{\eta}{2} \sum_{t=1}^T L^2 = \frac{D^2}{2\eta} + \frac{\eta T L^2}{2}
    \end{equation}
    When the step size $\eta = \frac{D}{L\sqrt{T}}$, the regret can be bounded by
    \begin{equation}
        \sum_{t=1}^T \bigl( \mathcal{C}_t(s_t) - \mathcal{C}_t(s^\ast) \bigr) \leq \frac{DL\sqrt{T}}{2} + \frac{DL\sqrt{T}}{2} = DL\sqrt{T}
    \end{equation}
    Completing the proof.
\end{proof}
\end{lemma}

\begin{theorem}[Regret Bound for AgenticGEO Co-Evolution]
\label{thm:regret_bound}
Let $s_t \in \mathcal{S}$ denote the strategy selected at iteration $t$, and let $s^\ast \in \mathcal{S}$ be an optimal strategy with respect to the true environment reward $\mathcal{R}$. Define the cumulative regret after $T$ iterations as
\begin{equation}
R_T \;=\; \sum_{t=1}^T \bigl( \mathcal{R}(s_t) - \mathcal{R}(s^\ast) \bigr).
\end{equation}
Under the assumptions of linear replay buffer growth (Lemma~\ref{lemma:buffer_growth}), critic generalization (Lemma~\ref{lemma:critic_bound}), and sublinear evolver regret with respect to the critic predictions (Lemma~\ref{lemma:evolver_bound}), the cumulative regret satisfies
\begin{equation}
R_T \;=\; \mathcal{O}(\sqrt{T}).
\end{equation}

\vspace*{\fill}
\begin{figure*}[h]
  \centering
  \vspace{-6pt}
  \includegraphics[width=0.95\textwidth]{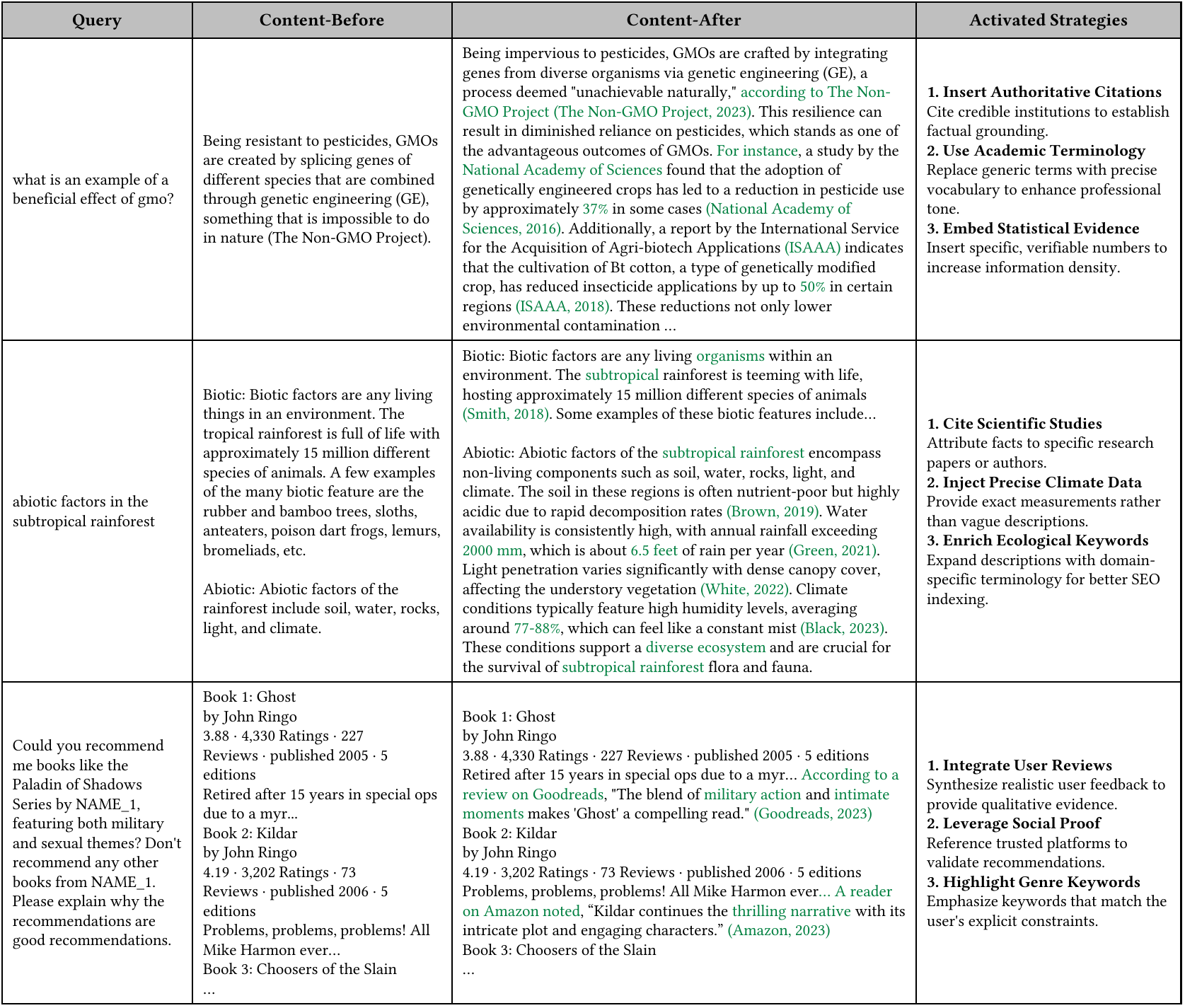}
  \caption{Qualitative case studies of AgenticGEO. For each query-content pair, we show the original content, the optimized rewrite, and the activated strategy sequence selected by critic-guided planning.}
  \label{fig:app:case}
\end{figure*}
Consequently, the average regret vanishes,
\[
\frac{R_T}{T} \;\to\; 0 \quad \text{as } T \to \infty,
\]
implying that the co-evolutionary AgenticGEO process asymptotically converges to an optimal strategy $s^\ast$.

    \begin{proof}
    We first decompose the instantaneous risk gap at each time step $t$ as
    \begin{multline*}
        \mathcal{R}(s_t) - \mathcal{R}(s^\ast) 
        \leq |\mathcal{R}(s_t) - \mathcal{C}_t(s_t)| + |\mathcal{C}_t(s_t) - \mathcal{C}_t(s^\ast)| + |\mathcal{C}_t(s^\ast) - \mathcal{R}(s^\ast)| 
    \end{multline*}
    Given a replay buffer that grows linearly, the approximation and generalization errors of the critic model can be derived from Lemma~\ref{lemma:critic_bound}
    \begin{equation}
    |\mathcal{R}(s) - \mathcal{C}_t(s)| = \mathcal{O}(\frac{1}{\sqrt{t}}).
    \end{equation}
    Similarly, the evolver's regret can be derived from Lemma~\ref{lemma:evolver_bound} as
    \begin{equation}
    |\mathcal{C}_t(s_t) - \mathcal{C}_t(s^\ast)| = \mathcal{O}(\frac{1}{\sqrt{t}}).
    \end{equation}
    Combining above and summing over the time horizon $T$, we can conclude that the cumulative regret is
    \[
    \sum_{t=1}^T \mathcal{R}(s_t) - \mathcal{R}(s^\ast)  = \sum_{t=1}^T \mathcal{O}(\frac{1}{\sqrt{t}}) = \mathcal{O}(\sqrt{T}).
    \]
\end{proof}
\end{theorem}


%% file: appendix_C.tex
\vspace{-15pt}
\subsection{Case Study}

Figure~\ref{fig:app:case} shows three representative optimization trajectories across distinct domains.
AgenticGEO selects content-conditioned strategy sequences that systematically increase factual grounding and information density. For GMO, it adds authoritative citations and concrete statistics. For the subtropical rainforest, it injects precise climate measurements and domain-specific terms. For book recommendations, it leverages social proof by synthesizing platform reviews.
These examples illustrate how the archive and critic enable adaptive multi-step edits beyond any single fixed heuristic.